\pdfoutput=1
\documentclass[10pt,twocolumn,letterpaper]{article}

\usepackage{cvpr}              

\usepackage[utf8]{inputenc} 
\usepackage[T1]{fontenc}    
\usepackage{graphicx}
\usepackage{booktabs} 
\usepackage{amssymb}
\usepackage{amsfonts}       
\usepackage{nicefrac}       
\usepackage{microtype}      
\usepackage{graphicx}
\usepackage{subcaption}
\makeatletter
\@namedef{ver@everyshi.sty}{}
\makeatother
\usepackage{tikz}
\usepackage{pgfplots}
\usepackage{amsmath}
\usepackage{threeparttable}
\usepackage{mathtools}
\usepackage{multirow}
\usepackage{soul}
\usepackage{amsthm}
\theoremstyle{plain}
\newtheorem*{theorem}{Complexity}
\usepackage[pagebackref]{hyperref}

\DeclarePairedDelimiterX{\infdivx}[2]{(}{)}{%
  #1\;\delimsize\|\;#2%
}
\newcommand{\infdiv}{D\infdivx}
\newcommand{\rulesep}{\unskip\ \vrule width 0.3ex\ }

\usepackage[linesnumbered,algoruled,vlined]{algorithm2e}

\SetAlCapNameFnt{\small}
\SetAlCapFnt{\small}
\SetAlgorithmName{Algorithm}{Algorithm}{List of Algorithms}

\pgfplotsset{compat=1.16}
\usepackage{environ}
\NewEnviron{myequation}{
    \begin{equation*}
    \scalebox{1}{$\BODY$}
    \end{equation*}
    }
\newcommand\Mark[1]{\textsuperscript{#1}}

\usepackage[capitalize]{cleveref}
\crefname{section}{Sec.}{Secs.}
\Crefname{section}{Section}{Sections}
\Crefname{table}{Table}{Tables}
\crefname{table}{Tab.}{Tabs.}

\def\cvprPaperID{\#7213} 
\def\confName{CVPR}
\def\confYear{2022}

\title{DO-GAN: A Double Oracle Framework for Generative Adversarial Networks
}
\author{
\textbf{Aye Phyu Phyu Aung\Mark{1,2}\thanks{Equal contribution}, Xinrun Wang\Mark{1}\footnotemark[1]~\thanks{
Corresponding author}, Runsheng Yu\Mark{1}\footnotemark[1], Bo An\Mark{1}\footnotemark[2]}\\ \textbf{Senthilnath Jayavelu\Mark{2}, Xiaoli Li\Mark{1,2,3}\footnotemark[2]} \\
\Mark{1}School of Computer Science and Engineering, Nanyang Technological University, Singapore \\ 
\Mark{2}Institute for Infocomm Research, A*STAR, Singapore\\
\Mark{3}A*STAR Centre for Frontier AI Research, Singapore\\
{\tt\small \{ayep0001,xinrun.wang, boan\}@ntu.edu.sg, runshengyu@gmail.com,}\\ {\tt\small \{j\_senthilnath, xlli\}@i2r.a-star.edu.sg}\\
}

\begin{document}
\maketitle

\begin{abstract}
In this paper, we propose a new approach to train Generative Adversarial Networks (GANs) where we deploy a double-oracle framework using the generator and discriminator oracles. GAN is essentially a two-player zero-sum game between the generator and the discriminator. Training GANs is challenging as a pure Nash equilibrium may not exist and even finding the mixed Nash equilibrium is difficult as GANs have a large-scale strategy space. In DO-GAN, we extend the double oracle framework to GANs. We first generalize the players' strategies as the trained models of generator and discriminator from the best response oracles. We then compute the meta-strategies using a linear program. For scalability of the framework where multiple generators and discriminator best responses are stored in the memory, we propose two solutions: 1) pruning the weakly-dominated players' strategies to keep the oracles from becoming intractable; 2) applying continual learning to retain the previous knowledge of the networks. We apply our framework to established GAN architectures such as vanilla GAN, Deep Convolutional GAN, Spectral Normalization GAN and Stacked GAN. Finally, we conduct experiments on MNIST, CIFAR-10 and CelebA datasets and show that DO-GAN variants have significant improvements in both subjective qualitative evaluation and quantitative metrics, compared with their respective GAN architectures.
\end{abstract}

\section{Introduction}
Generative Adversarial Networks (GANs)~\cite{goodfellow2014generative} have been applied in various domains such as image and video generation, text-to-image synthesis and equipment condition monitoring~\cite{liu2017unsupervised,reed2016generative,ragab2020adversarial,ragab2020contrastive}. Various architectures are proposed to generate more realistic samples~\cite{radford2015unsupervised,mirza2014conditional,pu2016variational} as well as regularization techniques~\cite{arjovsky2017wasserstein,miyato2018virtual}. From the game-theoretic perspective, GANs can be viewed as a two-player game where the generator samples the data and the discriminator classifies the data as real or generated. They are alternately trained to maximize their respective utilities till convergence corresponding to a pure Nash Equilibrium (NE). 
\begin{figure}[htbp]
\centering
\begin{subfigure}[b]{\linewidth}
\centering
\includegraphics[width=\linewidth]{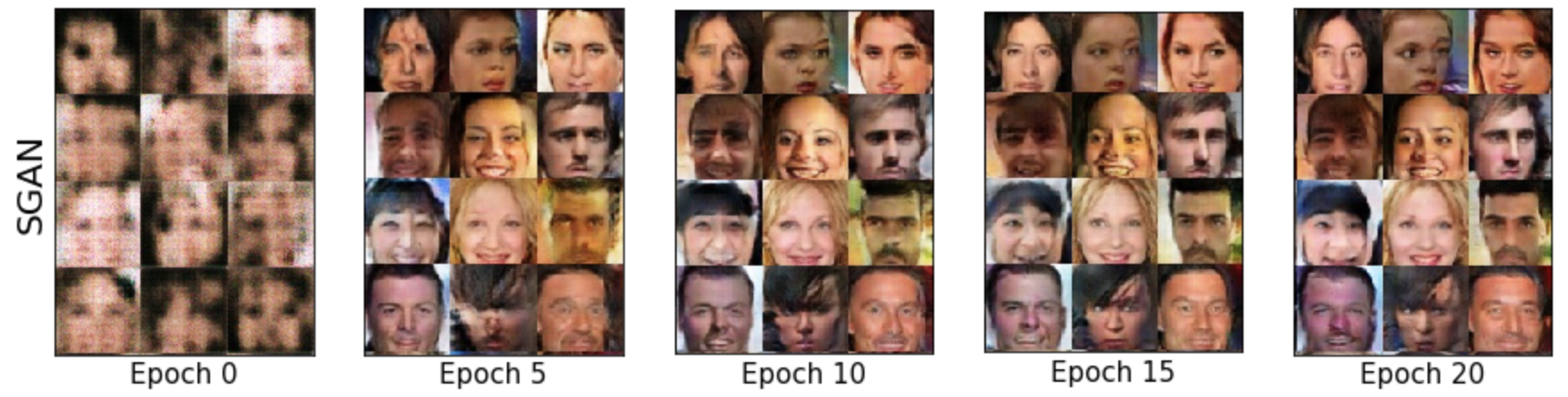}
\end{subfigure}
\begin{subfigure}[b]{\linewidth}
\centering
\includegraphics[width=\linewidth]{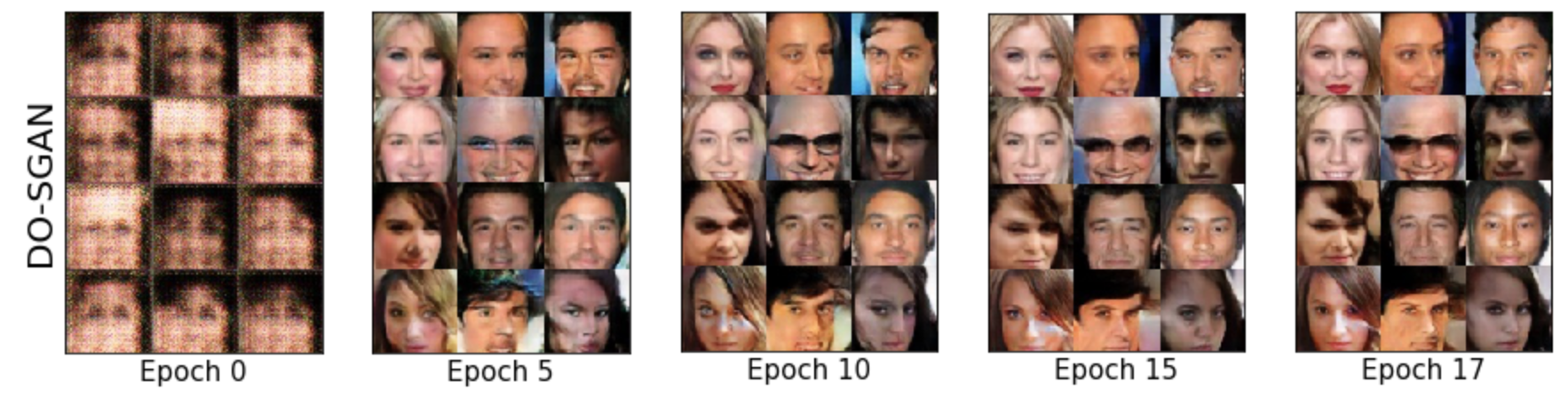}
\end{subfigure}
\caption{Training images with fixed noise for SGAN and DO-SGAN/P (pruning) until termination. Both during training and after convergence, DO-SGAN/P can generate better quality images with FID score of $6.32$ against SGAN's FID score of $6.98$.}
\label{sgandogsgan}
\end{figure}

However, pure NE cannot be reliably reached by existing algorithms as pure NE may not exist ~\cite{farnia2020gans,mescheder2017numerics}. This also leads to unstable training in GANs depending on the data and the hyperparameters. Therefore, mixed NE is a more suitable solution concept~\cite{hsieh2018finding}. Several recent works propose mixture architectures with multiple generators and discriminators that consider mixed NE such as MIX+GAN~\cite{arora2017generalization} and MGAN~\cite{hoang2018mgan} but they cannot guarantee to converge to mixed NE. Mirror-GAN~\cite{hsieh2018finding} computes the mixed NE by sampling over the infinite-dimensional strategy space and proposes provably convergent proximal methods. However, the sampling approach may not be efficient as mixed NE may only have a few strategies in the support set. 

Double Oracle (DO) algorithm~\cite{mcmahan2003planning} is a powerful framework to compute mixed NE in large-scale games.  
The algorithm starts with a restricted game that is initialized with a small set of actions and solves it to get the NE strategies of the restricted game. The algorithm then computes players' best-responses using oracles to the NE strategies and add them into the restricted game for the next iteration.
DO framework has been applied in various disciplines~\cite{jain2011double,bosansky2013double}, as well as Multi-agent Reinforcement Learning (MARL)~\cite{lanctot2017unified}.

Inspired by successful applications of DO framework, we, for the first time, propose a Double Oracle Framework for Generative Adversarial Networks (DO-GAN). This paper presents four key contributions. First, we treat the generator and the discriminator as players and obtain the best responses from their oracles and add the utilities to a meta-matrix. Second, we propose a linear program to obtain the probability distributions of the players' pure strategies (meta-strategies) for the respective oracles. The linear program computes an exact mixed NE of the meta-matrix game in polynomial time. Third, since multiple generators and discriminator from the best responses oracles are stored in the memory, the algorithm may be memory-inefficient for problems to train GAN with large-scaled real-world datasets. Thus, we propose two solutions for the scalable double oracle framework: 1) a pruning method for reducing the support set of best response strategies to prevent the oracles from becoming intractable as there is a risk of the meta-matrix growing very large with each iteration of oracle training; 2) applying continual learning to retain the previous knowledge of the networks for the best responses from the generator and discriminator oracles in the multi-task learning setup. We also address the problems in continual learning such as catastrophic forgetting. Finally, we provide comprehensive evaluation on the performance of DO-GAN with different GAN architectures using both synthetic and real-world datasets. Experiment results show that DO-GAN variants have significant improvements in terms of both subjective qualitative evaluation and quantitative metrics such as inception score and FID score.
\section{Related Works}
In this section, we briefly introduce existing GAN architectures, double oracle algorithm and its applications such as policy-state response oracles that are related to our work. 

\paragraph{GAN Architectures.}
Various GAN architectures have been proposed to improve the performance of GANs. Deep Convolutional GAN (DCGAN)~\cite{radford2015unsupervised} replaces fully-connected layers in the generator and the discriminator with deconvolution layer of Convolutional Neural Networks (CNN). Weight normalization techniques such as Spectral Normalization GAN (SNGAN)~\cite{miyato2018spectral} stabilize the training of the discriminator and reduce the intensive hyperparameters tuning. There are also multi-model architectures such as Stacked Generative Adversarial Networks (SGAN)~\cite{huang2017stacked} that consist of a top-down stack of generators and a bottom-up discriminator network. Each generator is trained to generate lower-level representations conditioned on higher-level representations that can fool the corresponding representation discriminator. Training GANs is very hard and unstable as pure NE for GANs might not exist and cannot be reliably reached by the existing approaches~\cite{mescheder2017numerics}. Considering mixed NE, MIX+GAN~\cite{arora2017generalization} maintains a mixture of generators and discriminators with the same network architecture but have their own trainable parameters. However, training a mixture of networks without parameter sharing makes the algorithm computationally expensive. Mixture Generative Adversarial Nets (MGAN)~\cite{hoang2018mgan} propose to capture diverse data modes by formulating GAN as a game between a classifier, a discriminator and multiple generators with parameter sharing. However, MIX+GAN and MGAN cannot converge to mixed NE. Mirror-GAN~\cite{hsieh2018finding} finds the mixed NE by sampling over the infinite-dimensional strategy space and proposes provably convergent proximal methods. The sampling approach may be inefficient to compute mixed NE as the mixed NE may only have a few strategies with positive probabilities in the infinite strategy space. 
 
\paragraph{Double Oracle Algorithm.}
Double Oracle (DO) algorithm starts with a small restricted game between two players and solves it to get the players' strategies at Nash Equilibrium (NE) of the restricted game. The algorithm then exploits the respective best response oracles for additional strategies of the players. The DO algorithm terminates when the best response utilities are not higher than the equilibrium utility of the current restricted game, hence, finding the NE of the game without enumerating the entire strategy space. Moreover, in two-player zero-sum games, DO converges to a min-max equilibrium~\cite{mcmahan2003planning}. DO framework is used to solve large-scale normal-form and extensive-form games such as security games~\cite{tsai2012security,jain2011double}, poker games~\cite{waugh2009strategy} and search games~\cite{bosansky2012iterative}. 
DO framework is also used in MARL settings~\cite{lanctot2017unified,muller2019generalized}. Policy-Space Response Oracles (PSRO) generalize the double oracle algorithm in a multi-agent reinforcement learning setting~\cite{lanctot2017unified}. PSRO treats the players' policies as the best responses from the agents' oracles, builds the meta-matrix game and computes the mixed NE but it uses Projected Replicator Dynamics (PRD) that updates the changes in the probability of each player's policy at each iteration. Since PRD needs to simulate the update for several iterations, the use of PRD takes a longer time to compute the meta-strategies and does not guarantee to compute an exact NE of the meta-matrix game. However, in DO-GAN, we can use a linear program to compute the players' meta-strategies in polynomial time since GAN is a two-player zero-sum game~\cite{schrijver1998theory}. We present the corresponding terminologies between GAN and game theory in Appendix~\ref{appendixTerminolgies}.

\paragraph{Continual Learning and Catastrophic Forgetting.}
 Continual learning in GANs has been ongoing research to combine a network’s knowledge through time or knowledge of multiple networks to a single network. Continual Learning in GANs \cite{seff2017continual} employed Elastic Weight Consolidation (EWC) to remedy the catastrophic forgetting in GANs continual training. MGAN~\cite{hoang2018mgan} and GMAN~\cite{durugkar2016generative} have employed continual learning to multiple generators and multiple discriminators respectively. Our work is closely related to Bayesian GAN \cite{saatchi2017bayesian} which assigns a posterior over the multiple networks of generator and discriminator. However, we cannot directly adapt the work as it only assigns a distribution to multiple generators and discriminators with a Bayesian formula without a single continual network while we assign the distributions to the generator/discriminator tasks of a continual learning architecture by solving a meta-game. 
\section{Preliminaries}
In this section, we mathematically explain the preliminary works to effectively our DO-GAN approach.

\subsection{Generative Adversarial Networks}
Generative Adversarial Networks (GANs)~\cite{goodfellow2014generative} have become one of the dominant methods for fitting generative models to complicated real-life data. GANs are deep neural net architectures comprised of two neural networks trained in an adversarial manner to generate data that resembles a distribution. The first neural network, a generator $G$, is given some random distribution $p_{\mathbf{z}}(\mathbf{z})$ on the input noise $\mathbf{z}$ and a real data distribution $p_{data}(\mathbf{x})$ on training data $\mathbf{x}$. The generator is supposed to generate as close as possible to $p_{data}(\mathbf{x})$. The second neural network, a discriminator $D$, is to discriminate between two different classes of data (real or fake) from the generator.

Let the generator's differentiable function be denoted as $G(\mathbf{z}, \pi_{g})$ and similarly $D(\mathbf{x}, \pi_{d})$ for the discriminator, where $G$ and $D$ are two neural networks with parameters $\pi_{g}$ and $\pi_{d}$. Thus, $D(\mathbf{x})$ represents the probability that $\mathbf{x}$ comes from the real data. The generator loss $L_{G}$ and the discriminator loss $L_{D}$ are defined as:
\begin{align}
\begin{split}
    L_{D} &= \mathbb{E}_{\mathbf{x} \sim p_{data}(\mathbf{x})}[-\log D(\mathbf{x})] \\ &+ \mathbb{E}_{\mathbf{z} \sim p_{\mathbf{z}}(\mathbf{z})}[-\log (1-D(G(\mathbf{z}))],
\end{split}\\
    L_{G} &= \mathbb{E}_{\mathbf{z} \sim p_{\mathbf{z}}(\mathbf{z})}[\log (1-D(G(\mathbf{z}))].
\end{align}

GAN is then set up as a two-player zero-sum game between $G$ and $D$ as follows:
\begin{multline}
    \min \nolimits_{G} \max \nolimits_{D} \thinspace \mathbb{E}_{\mathbf{x} \sim p_{data}(\mathbf{x})}[\log D(\mathbf{x})] \\ + \mathbb{E}_{\mathbf{z} \sim p_{\mathbf{z}}(\mathbf{z})}[\log (1-D(G(\mathbf{z}))].
    \label{zsg}
\end{multline}

During training, the parameters of $G$ and $D$ are updated alternately until we reach the global optimal solution $D(G(\mathbf{z}))= 0.5$. Next, we let $\Pi_g$ and $\Pi_{d}$ be the set of parameters for $G$ and $D$, considering the probability distributions $\sigma_{g}$ and $\sigma_{d}$, the mixed strategy formulation~\cite{hsieh2018finding} is:
\begin{small}
\begin{multline}
    \min_{\sigma_{g}} \max_{\sigma_{d}}\text{ } \mathbb{E}_{\pi_{d} \sim \sigma_{d}} \mathbb{E}_{\mathbf{x} \sim p_{data}(\mathbf{x})}[\log D(\mathbf{x},\pi_{d})] \\ +  \mathbb{E}_{\pi_{d} \sim \sigma_{d}} \mathbb{E}_{\pi_{g} \sim \sigma_{g}} \mathbb{E}_{\mathbf{z} \sim p_{\mathbf{z}}(\mathbf{z})}[\log (1-D(G(\mathbf{z},\pi_{g}),\pi_{d})].
    \label{mzsg}
\end{multline}
\end{small}\normalsize
Similarly to GANs, DCGAN, SNGAN and SGAN can also be viewed as two-player zero-sum games with mixed strategies of the players. DCGAN modifies the vanilla GAN by replacing fully-connected layers with the convolutional layers. SGAN trains multiple generators and discriminators using the loss as a linear combination of 3 loss terms: adversarial loss, conditional loss and entropy loss.

\subsection{Double Oracle Algorithm}
A normal-form game is a tuple $(\Pi, U, n)$ where $n$ is the number of players, $\Pi = (\Pi_{1}, \dots , \Pi_{n})$ is the set of strategies for each player $i \in N,$ where $N = \{1, \dots , n\}$ and $U : \Pi \rightarrow R^{n}$ is a payoff table of utilities $R$ for each joint policy played by all players. Each player chooses the strategy to maximize own expected utility from $\Pi_{i}$, or by sampling from a distribution over the set of strategies $\sigma_{i} \in \Delta(\Pi_{i})$. We can use linear programming, fictitious play~\cite{berger2007brown} or regret minimization~\cite{roughgarden2010algorithmic} to compute the probability distribution over players' strategies.

In the Double Oracle (DO) algorithm~\cite{mcmahan2003planning}, there are two best response oracles for the row and column player respectively. The algorithm creates restricted games from a subset of strategies at the point of each iteration $t$ for row and column players, i.e., $\Pi^{t}_{r} \subset \Pi_{r}$ and $\Pi^{t}_{c} \subset \Pi_{c}$ as well as a meta-matrix $U^{t}$ at the $t^{th}$ iteration. We then solve the meta-matrix to get the probability distributions on $\Pi^{t}_{r}$ and $\Pi^{t}_{c}$. 
Given a probability distribution $\sigma_{c}$ of the column player strategies, $BR_{r}(\sigma_{c})$ gives the row player's best response to $\sigma_{c}$. Similarly, given probability distribution $\sigma_{r}$ of the row player's strategies, $BR_{c}(\sigma_{r})$ is the column player's best response to $\sigma_{r}$. The best responses are added to the restricted game for the next iteration. The algorithm terminates when the best response utilities are not higher than the equilibrium utility of current restricted game. Although in the worst-case, the entire strategy space may be added to the restricted game, DO is guaranteed to converge to mixed NE in two-player zero-sum games. DO is also extended to the multi-agent reinforcement learning in PSRO~\cite{lanctot2017unified} to approximate the best responses to the mixtures of agents' policies, and compute the meta-strategies for the policy selection.

\section{DO-GAN}

As discussed in previous sections, computing mixed NE for GANs is challenging as there is an extremely large number of pure strategies, i.e., possible parameter settings of the generator and discriminator networks. Thus, we propose a double oracle framework for GANs (DO-GAN) to compute the mixed NE efficiently. DO-GAN builds a restricted meta-matrix game between the two players and computes the mixed NE of the meta-matrix game, then DO-GAN iteratively adds more generators and discriminators into the meta-matrix game until termination.

\subsection{General Framework of DO-GAN}
\begin{figure*}[ht]
    \centering
    \includegraphics[width=0.8\linewidth]{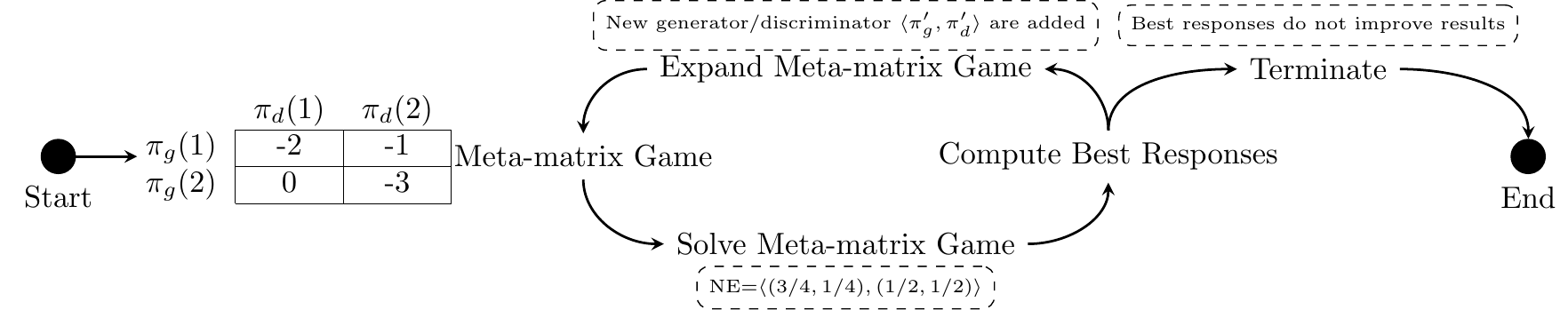}
    \caption{An illustration of DO-GAN. Figure adapted from~\cite{lanctot2017unified}.}
    \label{fig:illustrative_do_gan}
\end{figure*}

GAN can be translated as a two-player zero-sum game between the generator player $g$ and the discriminator player $d$. To compute the mixed NE of GANs, at iteration $t$, DO-GAN creates a restricted meta-matrix game $U^{t}$ with the trained generators and discriminators as strategies of the two players, where the generators and discriminators are parameterized by $\pi_{g}\in\mathcal{G}$ and $\pi_{d}\in\mathcal{D}$. We use $U^{t}(\pi_{g}, \pi_{d})$ to denote the generator player's payoff when playing $\pi_{g}$ against $\pi_{d}$, which is defined as $L_{D}$. Since GAN is zero-sum, the discriminator player's payoff is $-U^{t}(\pi_{g}, \pi_{d})$. We define $\sigma^{t}_{g}$ and $\sigma^{t}_{d}$ as the mixed strategies of generator player and discriminator player, respectively. With a slight abuse of notation, we define the generator player's expected utility of mixed strategies $\langle\sigma^{t}_{g},\sigma^{t}_{d}\rangle$ as $U^{t}(\sigma^{t}_{g}, \sigma^{t}_{d}) = \sum_{\pi_{g} \in \mathcal{G}} \sum_{\pi_{d} \in \mathcal{D}} \sigma^{t}_g(\pi_{g}) \cdot \sigma^{t}_{d}(\pi_{d}) \cdot U^{t}(\pi_{g},\pi_{d})$. We use $\langle\sigma_{g}^{t*},\sigma_{d}^{t*}\rangle$ to denote mixed NE of the restricted meta-matrix game $U^{t}$. We solve $U^{t}$ to obtain the mixed NE, compute best responses and add them into $U^{t}$ for next iteration. Figure~\ref{fig:illustrative_do_gan} presents an illustration of DO-GAN and Algorithm~\ref{dogan} describes the overview of the framework. 

Our algorithm starts by initializing two arrays $\mathcal{G}$ and $\mathcal{D}$ to store multiple generators and discriminators (line~\ref{initializeGD}). We train the first $\pi_{g}$ and $\pi_{d}$ with the canonical training procedure of GANs (line~\ref{trainCoreStart}). We store the parameters of trained models in $\mathcal{G}$ and $\mathcal{D}$ (line~\ref{assignmodels}), compute the adversarial loss $L_{D}$ and add it to the meta-matrix $U^{0}$ (line~\ref{computepayofffirst}). We initialize the meta-strategies $\sigma^{0*}_g=[1]$ and $\sigma^{0*}_d=[1]$ since there is only one pair of generator and discriminator available (line~\ref{initializeSigma}). For each epoch, we use $\mathtt{generatorOracle()}$ and $\mathtt{discriminatorOracle()}$ to obtain best responses $\pi'_{g}$ and $\pi'_{d}$ to $\sigma^{t*}_{d}$ and $\sigma^{t*}_{g}$ via Adam Optimizer, respectively, then add them into $\mathcal{G}$ and $\mathcal{D}$ (lines~\ref{bestresponseGStart}-\ref{bestresponseDEnd}). We then augment $U^{t-1}$ by adding $\pi'_{g}$ and $\pi'_{d}$ and calculating $U^{t}(\pi'_{g}, \pi'_{d})$ to obtain $U^{t}$ and compute the missing entries (line~\ref{payoffCompute}). We compute the missing payoff entries $U^{t}(\pi'_{g}, \pi_{d}), \forall \pi_{d} \in \mathcal{D}$ and $U^{t}(\pi_{g}, \pi'_{d}), \forall \pi_{g} \in \mathcal{G}$ by sampling a few batches of training data. After that, we compute the mixed NE $\langle\sigma_{g}^{t*},\sigma_{d}^{t*}\rangle$ of $U^{t}$ with linear programming (line~\ref{lp}). The algorithm terminates if the criteria described in Algorithm~\ref{term} is satisfied (line~\ref{terminationCheck}). 


In $\mathtt{generatorOracle}()$, we train $\pi'_{g}$ to obtain the best response against $\sigma^{t*}_{d}$, i.e., $U^{t}(\pi'_{g}, \sigma^{t*}_{d}) \geq U^{t}(\pi_{g}, \sigma^{t*}_{d}), \forall \pi_{g} \in \Pi_{g}$. Similarly, in $\mathtt{disciminatorOracle}()$, we train $\pi'_{d}$ to obtain the best response against $\sigma^{t*}_{g}$, i.e., $U^{t}(\sigma^{t*}_{g}, \pi'_{d}) \geq U^{t}(\sigma^{t*}_{g}, \pi_{d}), \forall \pi_d \in \Pi_{d}$. Full details of generator oracle and discriminator oracle can be found in Appendix~\ref{appendixA}.

\begin{algorithm}
\small
\DontPrintSemicolon
\caption{$\mathtt{DO-GAN()}$}
\label{dogan}
\SetKwInOut{Input}{input}
\SetKwComment{Comment}{//\ }{}
 \newcommand\mycommfont[1]{\footnotesize{#1}}
\SetCommentSty{mycommfont}
  Initialize generator and discriminator arrays $\mathcal{G} = \emptyset$ and $\mathcal{D} = \emptyset$; \label{initializeGD}\par
    Train generator \& discriminator to get the first $\pi_{g}$ and $\pi_{d}$; \label{trainCoreStart}   \par
  $\mathcal{G} \leftarrow \mathcal{G} \cup \{\pi_{g}\}$; $\mathcal{D} \leftarrow \mathcal{D} \cup \{\pi_{d}\}$; \label{assignmodels}\par
  Compute the adversarial loss $L_{D}$ and add it to meta-matrix $U^{0}$; \label{computepayofffirst}\par
  Initialize $\sigma^{0*}_g =[1]$ and $\sigma^{0*}_d=[1]$; \label{initializeSigma}\par
  \For{epoch $t \in \{1,2,...\}$}{
         $\pi'_{g} \leftarrow \mathtt{generatorOracle}(\sigma_{d}^{t*}, \mathcal{D})$; \label{bestresponseGStart}\par
        $\mathcal{G} \leftarrow \mathcal{G} \cup \{\pi'_{g}\}$; \label{bestresponseGEnd}\par
         $\pi'_{d} \leftarrow \mathtt{discriminatorOracle}(\sigma_{g}^{t*}, \mathcal{G})$;\label{bestresponseDStart} \par
        $\mathcal{D} \leftarrow \mathcal{D} \cup \{\pi'_{d}\}$; \par \label{bestresponseDEnd}\par
        Augment $U^{t-1}$ with $\pi'_{g}$ and $\pi'_{d}$ to obtain $U^{t}$ and compute missing entries; \label{payoffCompute} \par
       Compute mixed NE $\langle\sigma^{t*}_g, \sigma^{t*}_d\rangle$ for $U^{t}$ with linear program; \label{lp} \Comment*[r]{Section~\ref{lpMM}}
         \lIf(\hfill// \small{Section~\ref{termcheck}}){\texttt{\emph{TerminationCheck($U^{t}, \sigma^{t*}_g, \sigma^{t*}_d$)}
             \label{terminationCheck}}}{ 
            \textbf{break};
         } 
      }
\end{algorithm}

\subsection{Linear Program for Meta-matrix Game} \label{lpMM}
Since the current restricted meta-matrix game $U^{t}$ is a zero-sum game, we can use a linear program to compute the mixed NE in polynomial time~\cite{schrijver1998theory}.
Given the generator player $g$'s mixed strategy $\sigma^{t}_{g}$, the discriminator player $d$ will play strategies that minimize the expected utility of $g$. Thus, the mixed NE strategy for the generator player $\sigma^{t*}_{g}$ is to maximize the worst-case expected utility, which is obtained by solving the following linear program: 
\begin{multline}
    \sigma^{t*}_{g} = \arg \max\nolimits_{\sigma^{t}_{g}} \{v: \sigma^{t}_{g} \geq 0, \sum\nolimits_{i \in \mathcal{G}} \sigma^{t}_{g}(i) = 1, \\ U^{t}(\sigma^{t}_{g}, \pi_{d}) \geq v, \forall \pi_{d} \in \mathcal{D}\}.
    \label{lpeq}
\end{multline}
Similarly, we can obtain the mixed NE strategy for the discriminator $\sigma_{d}^{t*}$ by solving a linear program that maximizes the worst-case expected utility of the discriminator player. Therefore, we obtain the mixed NE $\langle\sigma^{t*}_{g}, \sigma^{t*}_{d}\rangle$ of the restricted meta-matrix game $U^{t}$.

\subsection{Termination Check}\label{termcheck}
DO terminates the training by checking whether the best response $\pi'_{g}$ (or $\pi'_{d}$) is in the support set $\mathcal{G}$ (or $\mathcal{D}$)~\cite{jain2011double}, but we cannot apply this approach to DO-GAN as GAN has infinite-dimensional strategy space~\cite{hsieh2018finding}.
Hence, we terminate the training if the best responses cannot bring a higher utility to the two players than the entries of the current support sets, as discussed in~\cite{lanctot2017unified,muller2019generalized}. Specifically, we first compute $U^{t}(\sigma^{t*}_{g}, \sigma^{t*}_{d})$ and the expected utilities for new generator and discriminator $U^{t}(\mathcal{G}[m], \sigma^{t*}_{d}), U^{t}(\sigma^{t*}_{g}, \mathcal{D}[n])$ (line~\ref{findUstart}-\ref{findUend}). Then, we calculate the utility increment (lines~\ref{gendiff}-\ref{disDiff}) and returns \textbf{True} if both $U^{t}(\mathcal{G}[m], \sigma^{t*}_{d})$ and $ U^{t}(\sigma^{t*}_{g}, \mathcal{D}[n])$ cannot bring a higher utility than $U^{t}(\sigma^{t*}_{g},\sigma^{t*}_{d})$ by $\epsilon$ (lines~\ref{diffStart}-\ref{diffEnd}).

\begin{algorithm}
\small
\caption{$\mathtt{Termination Check}(U^{t}, \sigma^{t*}_{g}, \sigma^{t*}_{d})$}
 \label{term}
\SetKwInOut{Input}{input}
\SetKwComment{Comment}{//\ }{}
 \newcommand\mycommfont[1]{\footnotesize{#1}}
\SetCommentSty{mycommfont}
\Comment{$U^{t}$ is of size $m \times n$}
\Comment{$|G|=m, |D|=n$}
Compute $U^{t}(\sigma^{t*}_{g},\sigma^{t*}_{d})$; \label{findUstart} \par
Compute $U^{t}(\sigma^{t*}_{g}, \mathcal{D}[n])$; \par
Compute $U^{t}(\mathcal{G}[m], \sigma^{t*}_{d})$; \label{findUend} \par
$\textit{genInc} = U^{t}(\mathcal{G}[m], \sigma^{t*}_{d}) - U^{t}(\sigma^{t*}_{g}, \sigma^{t*}_{d})$;\par \label{gendiff}
$\textit{disInc} = -U^{t}(\sigma^{t*}_{g}, \mathcal{D}[n]) -(- U^{t}(\sigma^{t*}_{g}, \sigma^{t*}_{d}))$; \par  \label{disDiff}
    \If{$\text{genInc} < \epsilon~\&\&~-\text{disInc} < \epsilon$  \label{diffStart}}{ 
    \Return\textbf{True}
}
\lElse{
    \Return\textbf{False} 
} \label{diffEnd}
\end{algorithm}

\begin{algorithm}
\small
\caption{DO-GAN/P}
 \label{prune}
\SetKwInOut{Input}{input}
\SetKwComment{Comment}{//\ }{}
 \newcommand\mycommfont[1]{\footnotesize{#1}}
\SetCommentSty{mycommfont}\
\Comment{I stores indices to be pruned from $\mathcal{G}$ and $\mathcal{D}$}
\Comment{G stores models to be pruned from $\mathcal{G}$ and $\mathcal{D}$}
$\mathtt{DO-GAN()}$;\par
$I_{g}=\emptyset; I_{d} = \emptyset$ \par
$K_{g}=\emptyset; K_{d}=\emptyset$ 
\If{$|\mathcal{G}| > s$ \label{pruneStart}}{
    \For{$i \in \{0, \dots, |\mathcal{G}|-1\}$}{
        \lIf{$\sigma^{t*}_{g}(\mathcal{G}[i]) == \min \sigma^{t*}_{g}$}{
            $I_{g} \leftarrow I_{g} \cup \{i\}$;
            $K_{g} \leftarrow K_{g} \cup \{\mathcal{G}[i]\}$
        }
    }
    }
    \If{$|\mathcal{D}| > s$}{
    \For{$j \in \{0, \dots, |\mathcal{D}|-1\}$}{
        \lIf{$\sigma^{t*}_{d}(\mathcal{D}[j]) == \min \sigma^{t*}_{d}$}{
            $I_{d} \leftarrow I_{d} \cup \{j\} $; 
            $K_{d} \leftarrow K_{d} \cup \{\mathcal{D}[j]\}$
        }
    }
    } 
    $\mathcal{G} \leftarrow \mathcal{G} \setminus K_{g}$; 
    $\mathcal{D} \leftarrow \mathcal{D} \setminus K_{d}$; \label{pruneEnd}\par
    $U \leftarrow J_{I_{g}, m} \cdot U^{t} \cdot J^{T}_{I_{d},n}$ \label{pruneU}
\end{algorithm}

\begin{algorithm}
\small
\DontPrintSemicolon
\caption{DO-GAN/C}
\label{contdogan}
\SetKwInOut{Input}{input}
\SetKwComment{Comment}{//\ }{}
 \newcommand\mycommfont[1]{\footnotesize{#1}}
\SetCommentSty{mycommfont}
  Initialize generator and discriminator task arrays $\mathcal{G} = \emptyset$ and $\mathcal{D} = \emptyset$; \label{initializeGDcont}\par
    Train generator \& discriminator to get with the first task to get $\pi^{0}_{g}$ and $\pi^{0}_{d}$; \label{trainCoreStartcont}   \par
  $\mathcal{G} \leftarrow \mathcal{G} \cup \{\pi_{g}\}$; $\mathcal{D} \leftarrow \mathcal{D} \cup \{\pi_{d}\}$; \label{assignmodelscont}\par
  Compute the adversarial loss $L_{D}$ and add it to meta-matrix $U^{0}$; \label{computepayofffirstcont}\par
  Initialize $\sigma^{0*}_g =[1]$ and $\sigma^{0*}_d=[1]$; \label{initializeSigmacont}\par
  \For{epoch $t \in \{1,2,...\}$}{
        Create new tasks $\pi^{t}_{g}$ and $\pi^{t}_{d}$; \label{createTask}\par
         $\pi^{t}_{g} \leftarrow \mathtt{generatorOracle}(\sigma_{d}^{t*}, \mathcal{D})$; \label{bestresponseGStartcont}\par
        $\mathcal{G} \leftarrow \mathcal{G} \cup \{\pi^{t}_{g}\}$; \label{bestresponseGEndcont}\par
         $\pi^{t}_{d} \leftarrow \mathtt{discriminatorOracle}(\sigma_{g}^{t*}, \mathcal{G})$;\label{bestresponseDStartcont} \par
        $\mathcal{D} \leftarrow \mathcal{D} \cup \{\pi^{t}_{d}\}$; \par \label{bestresponseDEndcont}\par
        \If{$t \geq 2$}{$\mathcal{G}\setminus \{ \pi^{t-2}_{g} \}$ \text{ and } $\mathcal{D} \setminus \{ \pi^{t-2}_{d} \}$ ;\label{removeprevious}}
        Create $U^{t}$ with $\pi^{t-1}_{g}, \pi^{t}_{g}$ and $\pi^{t-1}_{d}, \pi^{t}_{d}$; \label{payoffComputecont} \par
       Compute mixed NE $\langle\sigma^{t*}_g, \sigma^{t*}_d\rangle$ for $U^{t}$ with linear program; \label{lpcont} \Comment*[r]{Section~\ref{lpMM}}
         \lIf(\hfill// \small{Section~\ref{termcheck}}){\texttt{\emph{TerminationCheck($U^{t}, \sigma^{t*}_g, \sigma^{t*}_d$)}
             \label{terminationCheckcont}}}{ 
            \textbf{break};
         } 
      }
\end{algorithm}
\section{Practical Implementations}
As the number of epochs grows during the training of DO, the number of networks and the size of the meta-matrix also grows. Hence, there is a risk that the support strategy set becomes very large and $\mathcal{G}$ and $\mathcal{D}$ become intractable. To make the algorithm practical and scalable, we propose two methods: DO-GAN with meta-matrix pruning (DO-GAN/P) and DO-GAN with continual learning (DO-GAN/C).
\subsection{Meta-matrix Pruning  (DO-GAN/P)}\label{pruneMM}
The first method is to prune the meta-matrix. Here, we adapt the greedy pruning algorithm, as depicted in Algorithm~\ref{prune}. When either $|\mathcal{G}|$ or $|\mathcal{D}|$ is greater than the limit of the support set size $s$, we prune at least one strategy with the least probability, which is the strategy that contributes the least to the player's winning.
Specifically, we define $J_{I,b}$ where $I$ is the set of row numbers to be removed, $b$ is the total rows of a matrix. To remove the $2^{nd}$ row of a matrix having $3$ rows, we define $I= \{1\}, b=3$ and $J_{\{1\},3} =\left( \begin{matrix} 1 & 0 & 0  \\ 0 & 0 & 1  \end{matrix} \right)$. If $|\mathcal{G}|>s$, at least one strategy with minimum probability is pruned from $\mathcal{G}$, similarly for $\mathcal{D}$ (lines~\ref{pruneStart}-\ref{pruneEnd}). Finally, we prune the meta-matrix using matrix multiplication (line~\ref{pruneU}). 

\subsection{Continual Learning (DO-GAN/C)}
\label{sec:do_gan_c}
Our ablation studies show that we still need a support set of at least $s=10$ for DO-GAN/P to converge and the time complexity grows as $s$ increases. Thus, we further reduce both time and space complexity by making the network retain the knowledge of previous networks so that the algorithm will converge with even smaller support set. Hence, we propose to adapt continual learning to consolidate the knowledge of multiple networks to a single network while setting $s=2$ to reduce the space complexity as much as possible. We treat each network as a task to train the adaptive continual learning network while having a distribution over the tasks to represent the player's strategies. 

To remedy the catastrophic forgetting i.e., all the generator tasks focus only towards fooling the newest discriminator, we adapt Elastic Weight Consolidation (EWC) method~\cite{li2020few}. We change the generator loss function from non-saturating to saturating and add a penalty function accordingly. Let $G$ be trained on task $t-1$ to have optimal parameters $\pi^{t-1}_{g}$, the Fisher information $F$ is:
\begin{equation}
F = \mathbb{E}_{\mathbf{z} \sim p(\mathbf{z})} \Bigg[\Big(\frac{\delta}{\delta \pi^{t-1}_{g}} \log D(G(\mathbf{z})|\pi^{t-1}_{g})\Big)^{2} \Bigg]
 \end{equation}
 After obtaining the Fisher information, we directly use $F$ as a regularization loss to penalize the weight change during the training. Hence, $G$'s loss function is augmented as:
 \begin{multline}
 L_{G} = \mathbb{E}_{\mathbf{z} \sim p_{\mathbf{z}}(\mathbf{z})}[-\log D(G(\mathbf{z}) \\ + \lambda \cdot \sigma^{t*}_{g} \cdot \sum \nolimits_{i} F_{i} (\pi_{i} - \pi^{t-1,i}_{g})^{2}
\end{multline}
where $\pi^{t-1}_{g}$ represents the parameters learned for task $t-1$, $i$ is the index of each parameter of the generator model, and $\lambda$ is the regularization weight.

Algorithm \ref{contdogan} describes the changes to $\mathtt{DO-GAN}$ algorithm with continual learning. Instead of arrays $\mathcal{G}$ and $\mathcal{D}$ in $\mathtt{DO-GAN}$, we initialize tasks arrays for generator and discriminator (line \ref{initializeGDcont}). At every epoch, we create new tasks and train the generator and discriminator networks to outperform the previous optimal parameters with the distribution at NE $\sigma_{d}^{t*}$ and $\sigma_{g}^{t*}$ respectively (lines \ref{createTask}-\ref{bestresponseDEndcont}). Then, we keep the previously and currently trained optimal parameters (line \ref{removeprevious}) to create meta-matrix, solve it to compute mixed-NE (lines \ref{payoffComputecont}-\ref{lpcont}). Finally, we perform the termination check (line \ref{terminationCheckcont}).

\begin{theorem}
Given the same architectures, the space complexity of DO-GAN is $\mathcal{O}(t)$ where $t$ is the number of epochs until convergence. In contrast, the space complexity of DO-GAN/P is $\mathcal{O}(s)$ where $s$ is the size of the support set. DO-GAN/C has the minimum storage complexity which is $\mathcal{O}(1)$.
\end{theorem}

In DO-GAN, we add a pair of generator and discriminator for every epoch of training. Thus, the space complexity of DO-GAN is $\mathcal{O}(t)$, making the algorithm memory-inefficient to train with real-world datasets where it needs a large number of epochs to converge. The space complexity of DO-GAN/P is $\mathcal{O}(s)$ since we prune the meta-matrix and the players' strategies if the $|\mathcal{G}|>s$ or $|\mathcal{D}|>s$ where $s$ is the limit of the support set size. In DO-GAN/C, we train a single adaptive network storing the optimal strategies only for the tasks created at $(t-1)^{th}$ and $t^{th}$ epochs. Thus, the space complexity of DO-GAN/C is kept at $\mathcal{O}(1)$.
\section{Experiments}
We conduct our experiments on a machine with Xeon(R) CPU E5-2683 v3@2.00GHz and $4 \times$ Tesla v100-PCIE-16GB running Ubuntu operating system. We evaluate DO-framework for established GAN architectures such as vanilla GAN~\cite{goodfellow2014generative}, DCGAN~\cite{radford2015unsupervised}, SNGAN~\cite{miyato2018spectral} and SGAN~\cite{huang2017stacked}. We adopt the parameter settings and criterion of the GAN architectures as published.  We set $s=10$ unless mentioned otherwise. We compute the mixed NE of the meta game with Nashpy.
According to the ablation studies by \cite{seff2017continual,li2020few}, we set $\lambda$ as $1000$ for MNIST, $5000$ for CIFAR-10 and $5 \times 10^{8}$ for CelebA. The evaluation details are shown in Appendix~\ref{appendixB}.

\begin{figure}[ht]
    \centering
    \includegraphics[width=\linewidth]{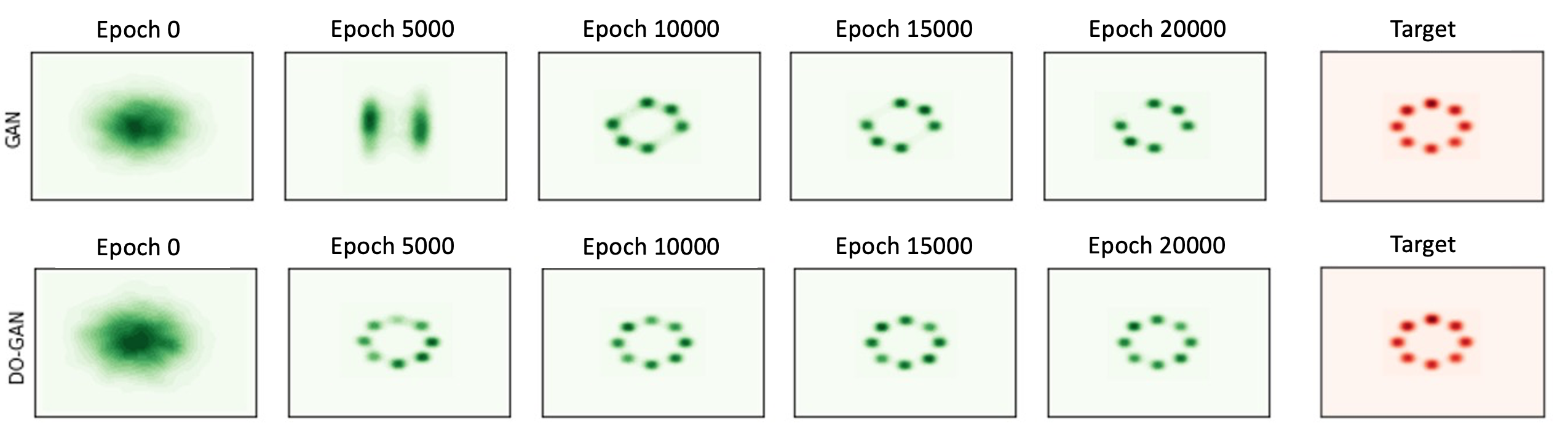}
    \caption{Comparison of GAN and DO-GAN/P on 2D synthetic Gaussian Mixture Dataset}
    \label{compare2d}
\end{figure}

\subsection{Evaluation on 2D Gaussian Mixture Dataset}
To illustrate the effectiveness of the architecture, we train a double oracle framework with the simple vanilla GAN architecture on a 2D mixture of 8 Gaussian mixture components with cluster standard deviation $0.1$ which follows the experiment by~\cite{metz2016unrolled}. Figure~\ref{compare2d} shows the evolution of 512 samples generated by GAN and DO-GAN/P through 20000 epochs. The goal of GAN and DO-GAN/P is to correctly generate samples at 8 modes as shown in the target. The results show that GAN can only identify 6 out of 8 modes of the synthetic Gaussian data distribution, while the DO-GAN/P can obtain all the 8 modes of the distribution. Furthermore, DO-GAN/P takes shorter time (less than 5000 epochs) to identify all 8 modes of the data distribution. We present a more detailed evolution of data samples through the training process on 2D Gaussian Mixtures in Appendix~\ref{appendixC}.

\paragraph{Ablations.} We also varied the limit of support set size for DO-GAN/P with $s=5,10,15$ and recorded the computation time as discussed in Appendix~\ref{appendixD}. We found that the training cannot converge when $s=5$ and takes a long time when $s=15$. Thus, we chose $s=10$ for the training.

\subsection{Evaluation on Real-world Datasets}
We evaluate the performance of the double oracle framework which takes several established GAN architectures as the backbone as discussed in Appendix~\ref{appendixG}, i.e., GAN, DCGAN and SGAN with convolutional layers for deep neural networks of GAN as well as SNGAN which uses normalization techniques. We run experiments on MNIST~\cite{lecun-mnisthandwrittendigit-2010}, CIFAR-10~\cite{cifarten} and CelebA~\cite{liu2015faceattributes}. MNIST contains 60,000 samples of handwritten digits with images of $28\times28$. CIFAR-10 contains $50,000$ training images of $32\times32$ of $10$ classes.
CelebA is a large-scaled face dataset with more than $200$K images of size $128\times128$.

\subsubsection{Qualitative Evaluation}
\begin{figure}[ht]
\centering
\includegraphics[width=\linewidth]{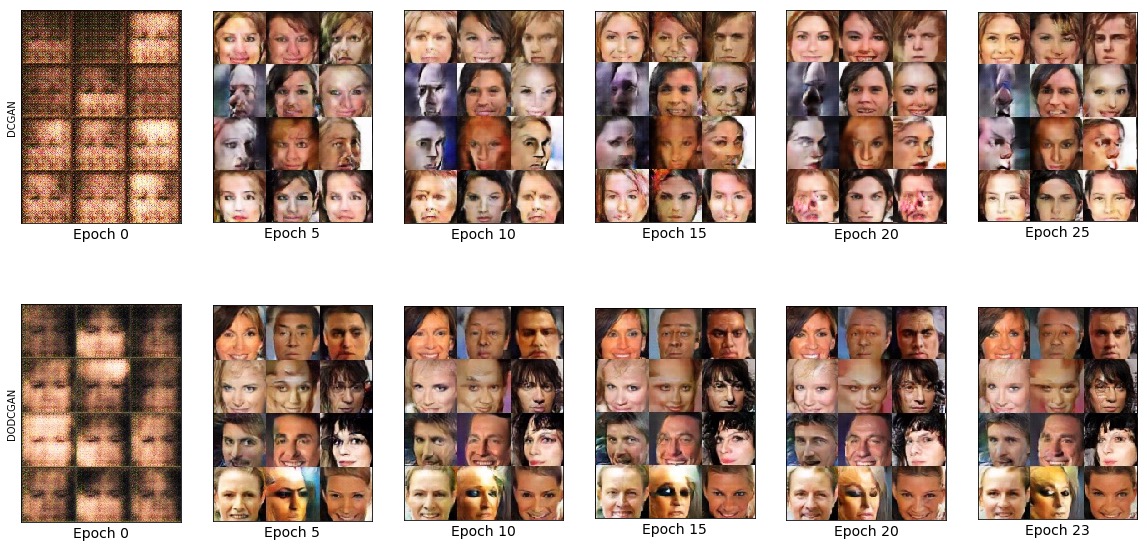}
\caption{Training images with fixed noise for DCGAN and DO-DCGAN/P until termination.}
\label{fig:celebmain}
\end{figure}

\begin{figure}[ht]
    \centering
    \includegraphics[width=\linewidth]{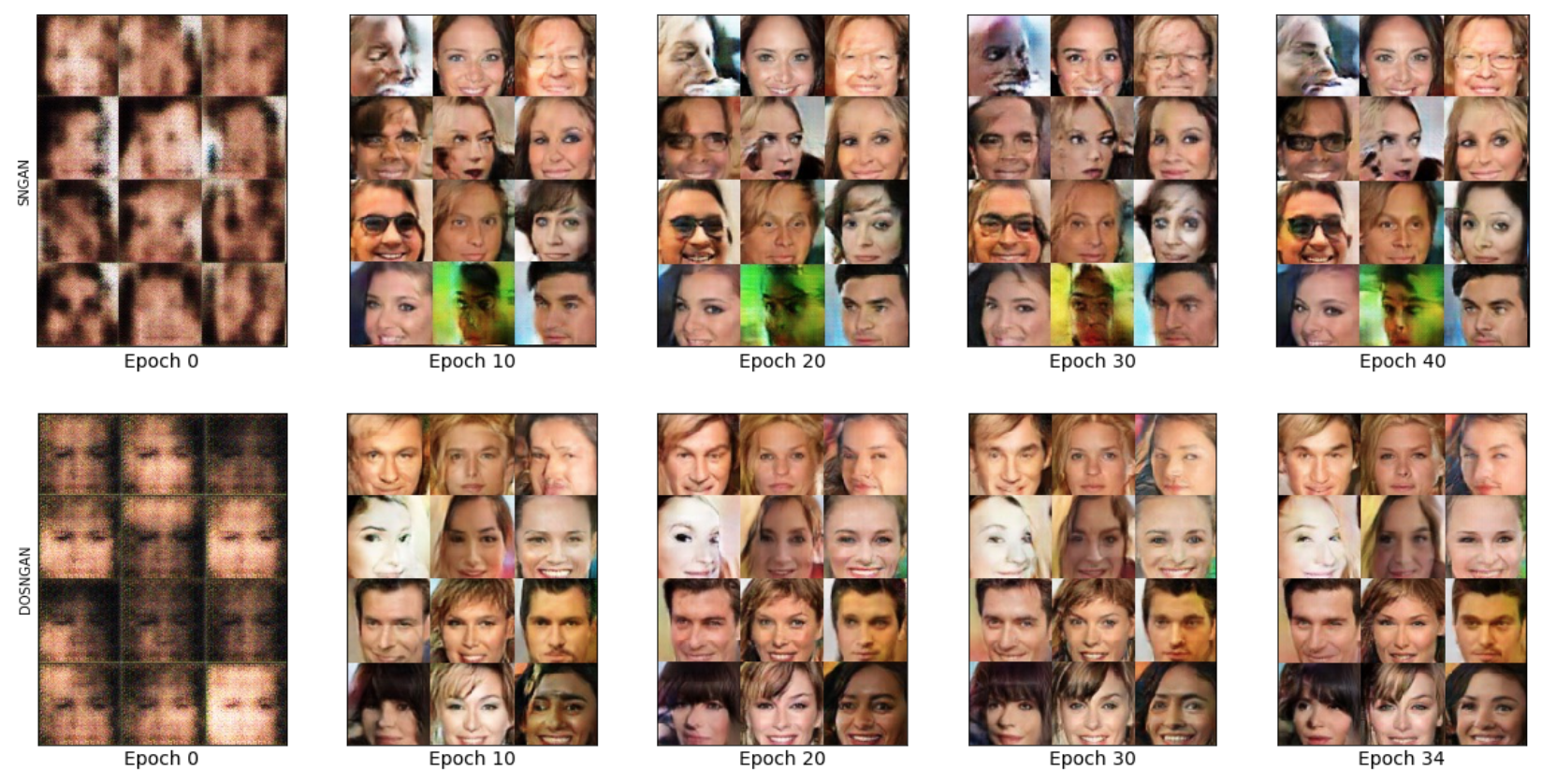}
    \caption{Training images with fixed noise for SNGAN and DO-SNGAN/P until termination.}
    \label{fig:celesnganmain}
\end{figure}

We choose the CelebA dataset for the qualitative evaluation since the training images contain noticeable artifacts (aliasing, compression, blur) that make the generator difficult to produce perfect and faithful images. We compare performances of DO-DCGAN/P, DO-SNGAN/P and DO-SGAN/P with their counterparts. SNGAN which is trained for 40 epochs with termination $\epsilon$ of $5 \times 10^{-5}$ for DO-SNGAN/P where other architectures are trained for 25 epochs with termination $\epsilon$ of $5 \times 10^{-5}$ for DO variants. The generated CelebA images of DCGAN and DO-DCGAN/P are shown in Figure~\ref{fig:celebmain}, where we find that DCGAN suffers
mode-collapse, while DO-DCGAN/P does not. We also present the generated images of SNGAN vs DO-SNGAN/P using fixed noise at different training epochs in Figure~\ref{fig:celesnganmain}. From the results, we can see that SNGAN, SGAN, DO-SNGAN/P and DO-SGAN/P are able to generate various faces, i.e., no mode-collapse. Judging from subjective visual quality, we find that DO-SNGAN/P and DO-SGAN/P are able to generate plausible images faster than SNGAN and SGAN during training, i.e., 17 epochs for DO-SGAN/P and 20 epochs for SGAN. More experimental results on CIFAR-10 and DO-GAN/C variants. which can produce competitive results, can be found in Appendix~\ref{appendixE}.

\subsubsection{Quantitative Evaluation}
In this section, we evaluate the performance of various architectures by quantitative metrics.
\paragraph{Inception Score.}
We first leverage the Inception Score (IS)~\cite{salimans2016improved} by using Inception\_v3~\cite{szegedy2016rethinking} as the inception model. To compute the inception score, we first compute the Kullback-Leibler (KL) divergence for all generated images and use the equation  $\text{IS} = \exp(\mathbb{E}_{\mathbf{x}} [\text{KL}(\infdiv{p(y|\mathbf{x})}{p(y)})])$ where $p(y)$ is the conditional label distributions for the images in the split and $p(y|\mathbf{x})$ is that of the image $\mathbf{x}$ estimated by the reference inception model. Inception score evaluates the quality and diversity of all generated images rather than the similarity to the real data from the test set. 

\paragraph{FID Score.}
Fr\'{e}chet Inception Distance (FID) measures the distance between the feature vectors of real and generated images using Inception\_v3 model~\cite{heusel2017gans}. Here, we let $p$ and $q$ be the distributions of the representations obtained by projecting real and generated samples to the last hidden layer of Inception model. Assuming that $p$ and $q$ are the multivariate Gaussian distributions, FID measures the 2-Wasserstein distance between the two distributions. Hence, FID Score can capture the similarity of generated images to real ones better than inception score.

\setuldepth{Berlin}

\begin{table*}[ht]
\caption{Inception scores (higher is better) and FID scores (lower is better) of DO-GAN with pruning (DO-GAN/P) and continual learning (DO-GAN/C). The mean and standard deviation are drawn from running 10 splits on 10000 generated images.}
\label{tab:score}
\centering
\begin{tabular}{@{}lllll@{}}
\toprule
         & \multicolumn{2}{c}{Inception Score}               & \multicolumn{2}{c}{FID Score} \\\cmidrule{2-5}
         & MNIST                  & CIFAR-10                 & CIFAR-10           & CelebA   \\ \midrule
GAN      & $1.04\pm0.05$          & $3.84\pm0.09$            & $71.44$            &   -     \\
DCGAN    & $1.26\pm0.05$          & $6.32 \pm 0.05$          & $37.66$            & $10.92$  \\
SNGAN    & $1.35\pm0.11$ & $7.58\pm0.12$ & $25.50$ & $7.62$ \\
SGAN     & \ul{$1.39\pm0.09$}          & \ul{$8.62\pm0.12$}   & \ul{$24.83$}            & \ul{$6.98$}   \\ \midrule
MIX+DCGAN &  - &  $7.72 \pm 0.09$     & - & -\\
MGAN    & - & $8.33\pm0.10$ & $26.70$  & - \\
DCGAN+ EWC & - & $7.58\pm0.07$ & $25.51$& -\\\midrule
DO-GAN/P   & $1.39\pm0.09$ (\textcolor{magenta}{$+0.35$})          & $7.20\pm0.16$ (\textcolor{magenta}{$+3.36$})            & $31.44$ (\textcolor{magenta}{$-40.00$})             &   -    \\
DO-DCGAN/P & $1.42\pm0.11$ (\textcolor{magenta}{$+0.16$})  & $7.86\pm 0.14$ (\textcolor{magenta}{$+1.54$})           & $22.25$ (\textcolor{magenta}{$-15.41$})             & $7.11$ (\textcolor{magenta}{$-3.81$})    \\ 
DO-SNGAN/P & $\mathbf{1.42\pm0.07}$ (\textcolor{magenta}{$+0.07$})  & $8.55\pm 0.08$ (\textcolor{magenta}{$+0.97$})  & $18.20$ (\textcolor{magenta}{$-7.30$})  & $6.92$ (\textcolor{magenta}{$-0.70$})  \\
DO-SGAN/P  & $1.42\pm0.09$ (\textcolor{magenta}{$+0.03$})  & $\mathbf{8.69 \pm 0.10}$ (\textcolor{magenta}{$+0.07$})  & $\mathbf{16.56}$ (\textcolor{magenta}{$-8.27$})   &  $\mathbf{6.32}$ (\textcolor{magenta}{$-0.66$})  \\ \midrule
DO-GAN/C   & $1.42\pm0.10$ (\textcolor{magenta}{$+0.38$})        &   $7.32\pm0.30$ (\textcolor{magenta}{$+3.48$})  &   $26.93$ (\textcolor{magenta}{$-44.51$})  &   -    \\
DO-DCGAN/C & $1.43\pm0.13$ (\textcolor{magenta}{$+0.17$})  & $8.04\pm0.22$ (\textcolor{magenta}{$+1.72$})         &  $21.50$ (\textcolor{magenta}{$-16.16$})       &  $7.16$ (\textcolor{magenta}{$-3.76$})  \\ 
DO-SNGAN/C & $1.43\pm0.08$ (\textcolor{magenta}{$+0.08$}) & $8.54\pm0.16$ (\textcolor{magenta}{$+0.96$}) & $19.57$ (\textcolor{magenta}{$-5.93$}) & $6.74$ (\textcolor{magenta}{$-0.88$}) \\
DO-SGAN/C  & $\mathbf{1.45\pm0.15}$ (\textcolor{magenta}{$+0.06$}) & $\mathbf{9.78\pm0.11}$ (\textcolor{magenta}{$+1.16$}) & $\mathbf{16.07}$ (\textcolor{magenta}{$-8.76$}) & $\mathbf{6.30}$ (\textcolor{magenta}{$-0.68$})  \\ \bottomrule
\end{tabular}
\begin{tablenotes}
    \centering
      \footnotesize
      \item Note: MIX+DCGAN and MGAN results are directly copied from~\cite{arora2017generalization,hoang2018mgan}. The magenta values are the improvements from non-DO counterparts.
    \end{tablenotes}
\end{table*}

\paragraph{Results.} The results are shown in Table~\ref{tab:score}.
In CIFAR-10 dataset, the pruning method i.e., DO-GAN/P, DO-DCGAN/P and DO-SNGAN/P obtain much better results ($7.2\pm 0.16$, $7.86\pm0.14$ and $8.55\pm0.08$) than GAN, DCGAN and SNGAN ($3.84 \pm 0.09$, $6.32\pm0.05$ and $7.58\pm0.12$). However, we do not see a significant improvement in DO-SGAN/P compared to SGAN $8.62 \pm 0.12$ and $8.69 \pm 0.10$ since SGAN already can generate diverse images. We did not include IS for CelebA dataset as IS cannot reflect the real image quality for CelebA, as observed in~\cite{heusel2017gans}. In CIFAR-10 dataset, DO-GAN/P, DO-DCGAN/P, DO-SNGAN/P and DO-SGAN/P obtain much lower FID scores ($31.44$, $22.25$, $16.56$, $18.20$) respectively. The trend follows in CelebA obtaining $7.11$ for DO-DCGAN/P while $10.92$ for DCGAN, $7.62$ for SNGAN while $6.92$ for DO-SNGAN/P, $6.98$ for SGAN and $6.32$ for DO-SGAN/P respectively. Although we see a significant improvement in the quality of DO-SGAN/P images, FID score for DO-SGAN/P is affected by distortions. 

We observe the competitive results for continual learning method with the pruning method:  DO-GAN/C, DO-DCGAN/C, DO-SNGAN/C and DO-SGAN/C obtain inception scores of $7.32\pm0.30$, $8.04\pm0.22$, $8.54\pm0.16$ and $9.78\pm0.11$ respectively as well as FID scores of $26.93$, $21.50$, $19.57$ and $16.07$ respectively for CIFAR-10 dataset. Moreover, $7.16$, $6.74$ and $6.30$ respectively for CelebA dataset. We also compared with DCGAN+EWC which uses continual learning without the double oracle framework for CIFAR-10 dataset and obtained better results where DCGAN+EWC obtained inception score of $7.58\pm0.07$ and FID score of $25.51$. 

\begin{table}[ht]
\caption{Wall-clock time comparison of SGAN variants}
\label{tab:wallclock}
\centering
\begin{tabular}{@{}llll@{}}
\toprule
         & SGAN                  & DO-SGAN/P               & DO-SGAN/C          \\ \midrule
Time (GPU-Hrs)      & $143.28$          & $119.04$            & $\mathbf{97.54}$   \\ \bottomrule
\end{tabular}
\begin{tablenotes}
    \centering
      \footnotesize
      \item Note: SGAN is trained for 500 epochs on CIFAR-10. DO-SGAN/P converged at 288 epochs and DO-SGAN/C at 236 epochs.
    \end{tablenotes}
\end{table}

From the results, we can see that DO framework performs better than each of their original counterpart architectures with both methods: pruning and continual learning. More details can be found in Appendix~\ref{appendixF}. We can also see that continual learning method that uses least storage space can obtain competitive results with DO-GAN/P without memory storage limitations or pruning the players' strategies. 

We also compare the wall-clock running times of DO-SGAN/P and DO-SGAN/C with SGAN as shown in Table~\ref{tab:wallclock}. We let SGAN train for 500 epochs on CIFAR-10 dataset. Meanwhile, DO-SGAN/P converged at 288 epochs and DO-SGAN/C converged at 236 epochs. The recorded GPU hours of 143.28 for SGAN, 119.04 for DO-SGAN/P and 97.54  for DO-SGAN/C show that DO variants are more efficient.
\section{Conclusion}
We propose a novel double oracle framework to GANs, which starts with a restricted game and incrementally adds the best responses of the generator and the discriminator oracles as the players' strategies. We then compute the mixed NE to get the players' meta-strategies by using a linear program. We also propose two approaches to make the solution scalable including pruning the support strategy set and continual learning with an adaptive architecture to store the multiple networks of generators and discriminators. We apply DO-GAN approach to established GAN architectures such as vanilla GAN, DCGAN, SNGAN and SGAN. Extensive experiments with the synthetic 2D Gaussian mixture dataset as well as real-world datasets such as MNIST, CIFAR-10 and CelebA show that DO-GAN variants have significant improvements in comparison to their respective GAN architectures in terms of both subjective image quality and quantitative metrics.

\newpage
\section*{Acknowledgement}
This research was supported by the National Research Foundation, Singapore under its AI Singapore Programme (AISG Award No: AISG-RP-2019-0013), National Satellite of Excellence in Trustworthy Software Systems (Award No: NSOE-TSS2019-01), and NTU.
{\small
\bibliographystyle{ieee_fullname}
\bibliography{egbib}
}

\newpage
\pagenumbering{roman}
\setcounter{page}{0}
\appendix
\onecolumn
\setcounter{figure}{0}
\setcounter{table}{0}
\setcounter{algocf}{0}
\section{Comparison of Terminologies between Game Theory and GAN}
\label{appendixTerminolgies}
\begin{table*}[ht]
\centering
\caption{Comparison of Terminologies between Game Theory and GAN}
\begin{tabular}{ll}
\toprule\toprule
Game Theory terminology & GAN terminology \\ \midrule\midrule
Player                  & Generator/ discriminator  \\ \midrule
Strategy                & The parameter setting of generator/ discriminator, e.g., $\pi_g$ and $\pi_d$ \\\midrule
\multirow{2}{*}{Policy}               & The sequence of parameters (strategies) till epoch $t$, e.g., ($\pi^1_g, \pi^2_g, ..., \pi^t_g$) \\
& Note: Not used in DO-GAN.\\\midrule
Game                    & The minmax game between generator and discriminator\\\midrule
\multirow{2}{*}{Meta-game/ meta-matrix}  & The minmax game between generator \& discriminator with \\
& their respective set of strategies at epoch $t$ of DO framework                             \\\midrule
Meta-strategy           & The mixed NE strategy of generator/discriminator at epoch $t$  \\ \bottomrule\bottomrule                                    
\end{tabular}
\end{table*}

\section{Full Algorithm of DO-GAN}
\label{appendixA}

\begin{algorithm}[ht]
\caption{\texttt{GeneratorOracle($\sigma^{t*}_{d}, \mathcal{D}$)}}
\SetKwInOut{Input}{input}
\SetKwComment{Comment}{//\ }{}
 \newcommand\mycommfont[1]{\footnotesize{#1}}
\SetCommentSty{mycommfont}
  Initialize a generator $G$ with random parameter setting $\pi'_{g}$; \par 
  \For{iteration $k_{0} \dots k_{n}$}{
    Sample noise $\mathbf{z}$; \par
    $\pi_{d} =$ Sample a discriminator from $\mathcal{D}$ with $\sigma^{t*}_d$; \par
    Initialize a discriminator $D$ with parameter setting $\pi_{d}$; \par
    Update the generator $G$'s parameters $\pi'_{g}$ via Adam optimizer:
    \begin{myequation}
        \bigtriangledown_{\pi'_{g}} \text{ log }(1 - D(G(\mathbf{z})))
    \end{myequation}
  }
\end{algorithm}

\begin{algorithm}[ht]
\caption{\texttt{DiscriminatorOracle($\sigma^{t*}_{g}, \mathcal{G}$)}}
\SetKwInOut{Input}{input}
\SetKwComment{Comment}{//\ }{}
 \newcommand\mycommfont[1]{\footnotesize{#1}}
\SetCommentSty{mycommfont}
  Initialize a discriminator $D$ with random parameter setting $\pi'_{d}$; \par 
  \For{iteration $k_{0} \dots k_{n}$}{
    Sample a minibatch of data $\mathbf{x}$; \par
    \For{a minibatch}{
    Sample noise $\mathbf{z}$; \par
    $\pi_{g} = $ Sample a generator from $\mathcal{G}$ with $\sigma^{t*}_g$; \par
    Initialize a generator $G$ with a parameter setting $\pi_{g}$; \par
    Generate and add to mixture $G(\mathbf{z})$;
    }
    Update the discriminator $D$'s parameters $\pi'_{d}$ via Adam optimizer: \par
    \begin{myequation}
        \bigtriangledown_{\pi'_{d}} \text{ log } D(\mathbf{x}) + \text{ log }(1 - D(G(\mathbf{z})))
    \end{myequation}
    
  }
\end{algorithm}

We train the oracles for some iterations which we denote as $k_{0,1,2, \dots}$. For experiments, we train each oracle for an epoch for the real-world datasets and $50$ iterations for the 2D Synthetic Gaussian Dataset. At each iteration $t$, we sample the generators from the support set $\mathcal{G}$ with the meta-strategy $\sigma^{t*}_{g}$ to generate the images for evaluation. Similarly, we conduct the performance evaluation with the generators sampled from $\mathcal{G}$ with the final $\sigma^{*}_{g}$ at termination. SGAN consists of a top-down stack of GANs, e.g, for a stack of 2, Generator 1 is the first layer stacked on Generator 0 with each of them connected to Discriminator 1 and 0 respectively. Hence, in DO-SGAN, we store the meta-strategies for the Generator 0 and 1 in $\sigma^{t*}_{g}$ and the Discriminator 1 and 0 for $\sigma^{t*}_{d}$. In $\mathtt{GeneratorOracle()}$, we first sample Discriminator 1 and 0 from discriminator distribution $\sigma^{t*}_{d}$ and train Generator 1 first then followed by calculating loss with Discriminator 1 and train Generator 0 subsequently, and finally calculate final loss with Discriminator 0 and train the whole model end to end. We perform the same process for $\mathtt{DisciminatorOracle()}$.
\newpage
\section{Implementation Details}
\label{appendixB}
\begin{table}[ht]
\caption{Training Hyperparameters}
\label{Hyper}
\centering
\begin{tabular}{lllll}
\toprule
                            & GAN    & DCGAN  & SNGAN & SGAN \\\midrule
Generator Learning Rate     & $0.0002$ & $0.0002$ & $0.0002$ & $0.0001$ \\
Discriminator Learning Rate & $0.0002$ & $0.0002$ & $0.0002$ & $0.0001$ \\
batch size                  & $64$     & $64$     & $64$ & $100$    \\
Adam: beta 1                & $0.5$   & $0.5$    & $0.5$ & $0.5$    \\
Adam: beta 2               & $0.999$  & $0.999$  & $0.999$ & $0.999$ \\\bottomrule
\end{tabular}
\end{table}

We implement our proposed method with Python 3.7, Pytorch=1.4.0 and Torchvision=0.5.0.  We set the hyperparameters as the original implementations. We present the hyperparameters set in Table~\ref{Hyper}. We use Nashpy to compute the equilibria of the meta-matrix game. 

\subsection{Value of $\lambda$}
The experiment in~\cite{li2020few} has done ablation studies for FFHQ dataset which are emoji faces and hence we used the $\lambda$ value for CelebA. Meanwhile, we adopted the results from~\cite{seff2017continual} and set 1000 for MNIST and the maximum value of ablation study for SVHN dataset to train CIFAR-10 as we want to use $\lambda$ values from the most similar datasets. The experiments in~\cite{seff2017continual} reported that they observed little difference in visual quality regarding with ablation study but high values of $\lambda$ cause no loss in visual fidelity when beginning training on a new task rather than lower value of $\lambda$. Hence, we use the maximum value.

\newpage
\section{Full Training Process of 2D Gaussian Dataset}
\label{appendixC}
\begin{figure}[ht]
\centering
  \begin{subfigure}[b]{0.45\linewidth}
    \includegraphics[width=\linewidth]{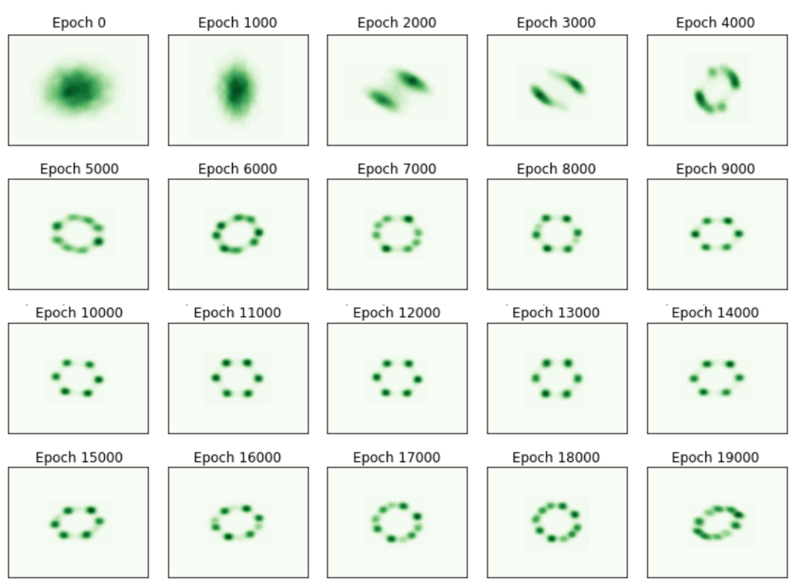}
    \caption{GAN}
  \end{subfigure}
  \hspace{5pt}
  \begin{subfigure}[b]{0.45\linewidth}
    \includegraphics[width=\linewidth]{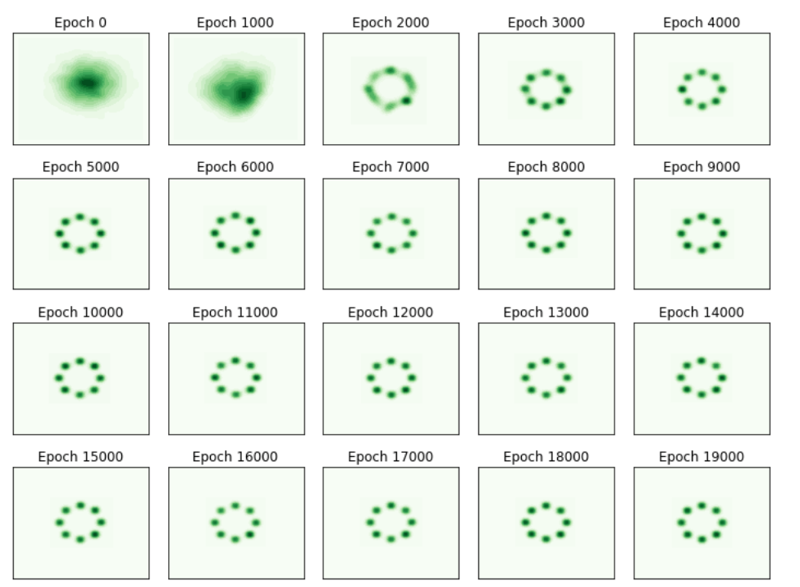}
    \caption{DO-GAN/P}
  \end{subfigure}
  \caption{Full comparison of GAN and DO-GAN/P on 2D Synthetic Gaussian Dataset}
  \label{fulltrain}
\end{figure}

\begin{figure}[ht]
    \centering
     \includegraphics[width=\linewidth]{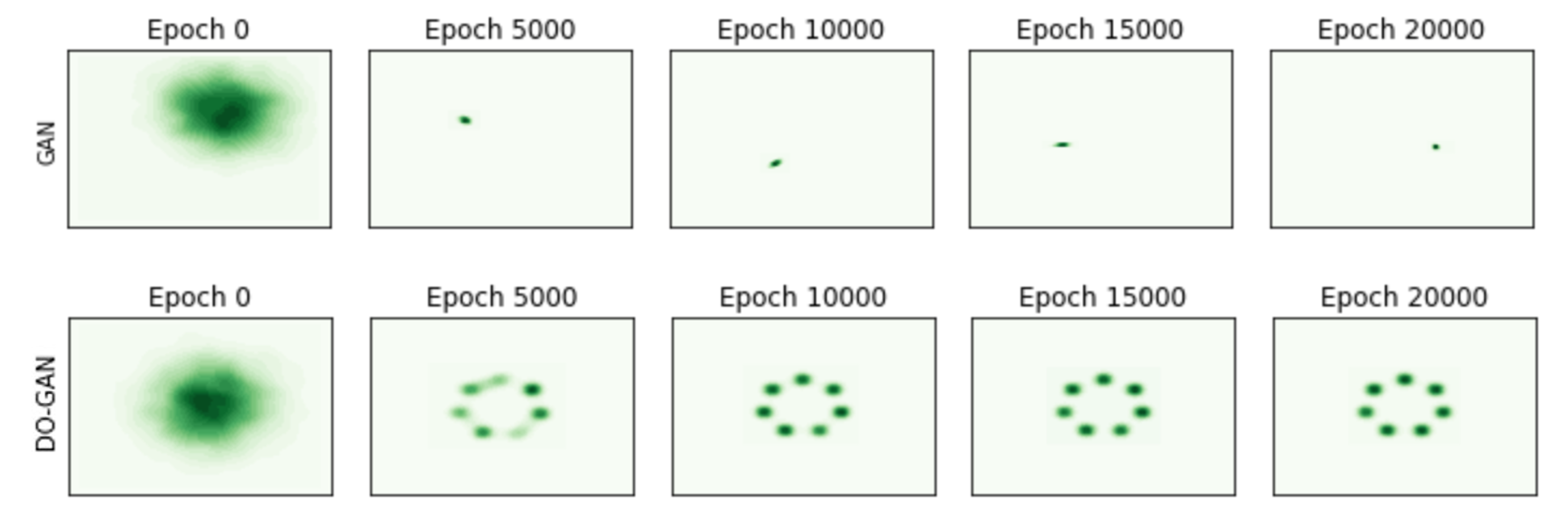}
     \caption{GAN and DO-GAN/P comparison with Gaussian Mixture 7 modes}
     \label{seven}
\end{figure}

\begin{figure}[ht]
    \centering
    \includegraphics[width=\linewidth]{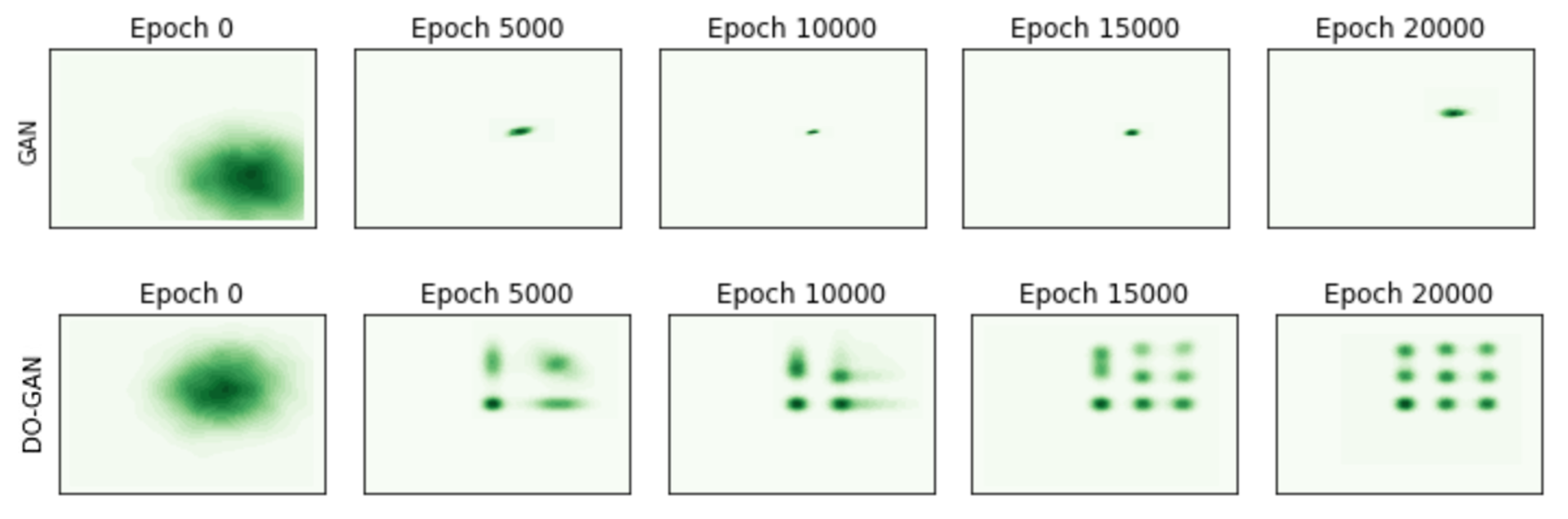}
    \caption{GAN and DO-GAN/P comparison with Gaussian Mixture 9 modes}
    \label{nine}
\end{figure}

Figure~\ref{fulltrain} shows the full training process of DO-GAN/P and GAN on 2D Synthetic Gaussian Dataset. From the results, we find that GAN struggles to generate the samples into $8$ modes while DO-GAN/P can generate all the 8 modes of the distribution. Furthermore, DO-GAN/P takes shorter time (less than $5000$ iterations) to identify all $8$ modes of the data distribution. Moreover, we present the experiment results on 7 mode and 9-mode Gaussian Mixtures in Figure~\ref{seven} and ~\ref{nine}.

\newpage

\section{Investigation of Support Set Size for DO-GAN/P}
\label{appendixD}
\begin{figure}[ht]
\centering
    \includegraphics[width=\linewidth]{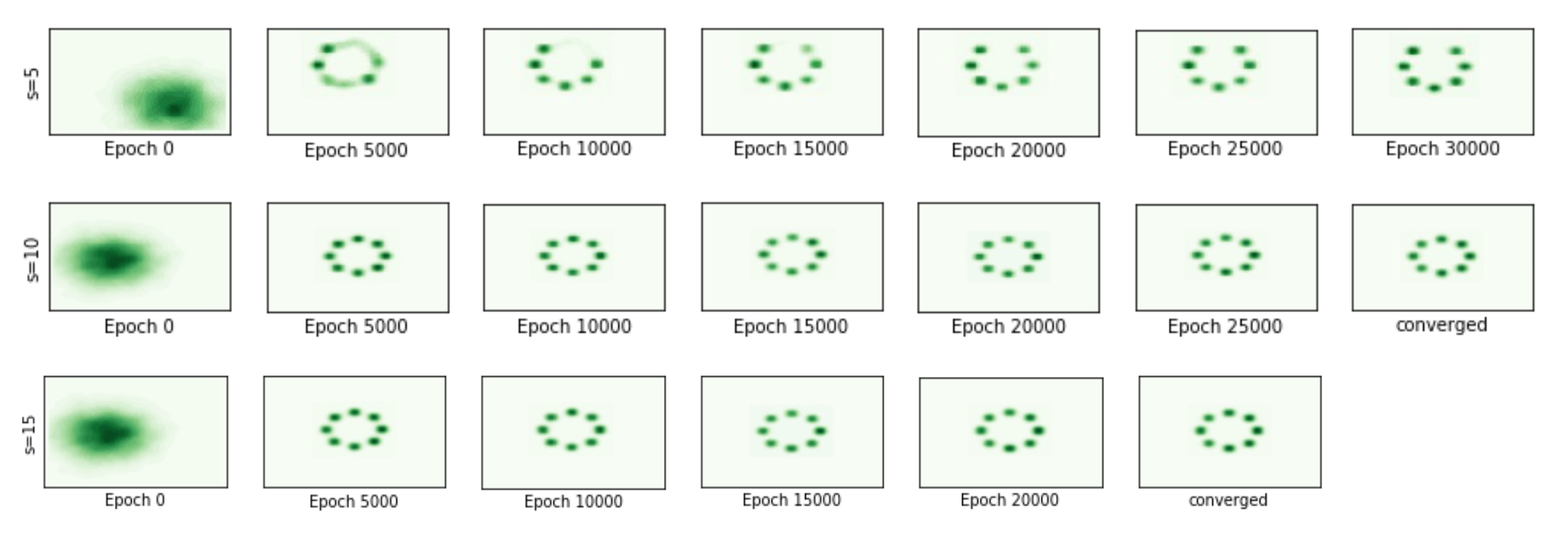}
  \caption{Training evolution on 2D Gaussian Dataset with $s=5,10,15$}
  \label{evolFig}
\end{figure}

We vary the support set size $s$ to $5,10,15$ and record the training evolution and the running time as presented in Table~\ref{runtimeTable} and Figure~\ref{evolFig}. We find that if the support size is too small, e.g., $s=5$, the best responses which are not optimal yet have better utilities than the models in the support set are added and pruned from the meta-matrix repeatedly making the training not able to converge.  However, $s=15$ takes a significantly longer time as the time for the augmenting of meta-matrix becomes exponentially long with the support set size. Hence, we chose $s=10$ as our experiment support set size since we observed that there is no significant trade-off and shorter runtime. 
\begin{table}[ht]
\centering
\caption{Runtime of DO-GAN/P on 2D Gaussian Dataset with $s=5,10,15$}
\label{runtimeTable}
\begin{tabular}{ll}
\toprule
Support Set Size & Runtime (GPU hours)                    \\ \midrule
$s=5$              & $> 1$   \\ 
$s=10$             & $0.5627$                                  \\ 
$s=15$             & $0.9989$                              \\ \bottomrule
\end{tabular}
\end{table}

\newpage
\section{Generated images of CelebA and CIFAR-10}
\label{appendixE}
In this section, we present the training images of CelebA and CIFAR-10 datasets. We do not evaluate the performance of vanilla GAN and its DO variant on CelebA dataset since DCGAN and SGAN outperform vanilla GAN in image generation tasks~\cite{radford2015unsupervised}.
\begin{figure}[!htb]
    \centering
    \includegraphics[width=\linewidth]{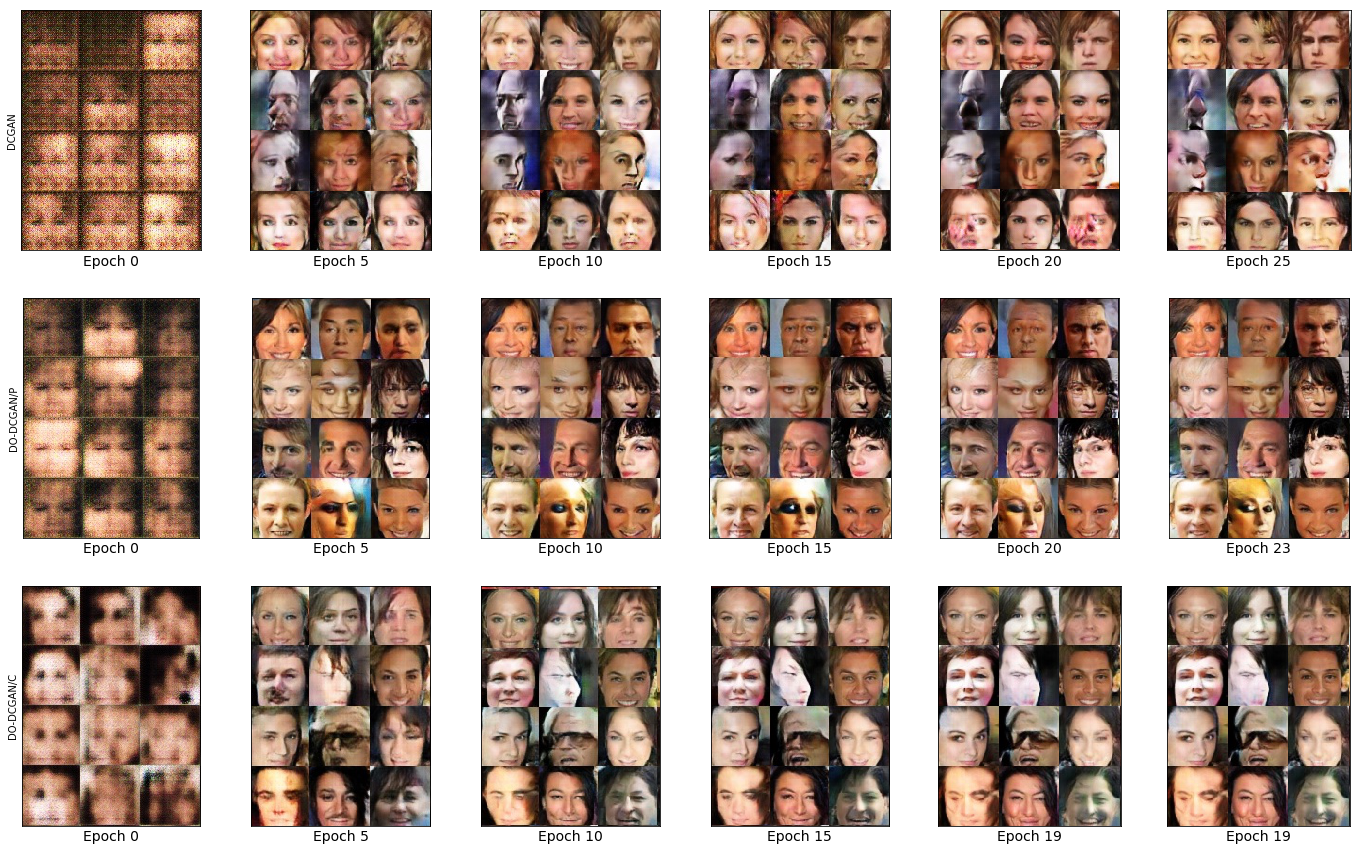}
    \caption{Training images with fixed noise for DCGAN and DO-DCGAN until termination.}
    \label{fig:celeb}
\end{figure}

\begin{figure}[!htb]
    \centering
    \includegraphics[width=\linewidth]{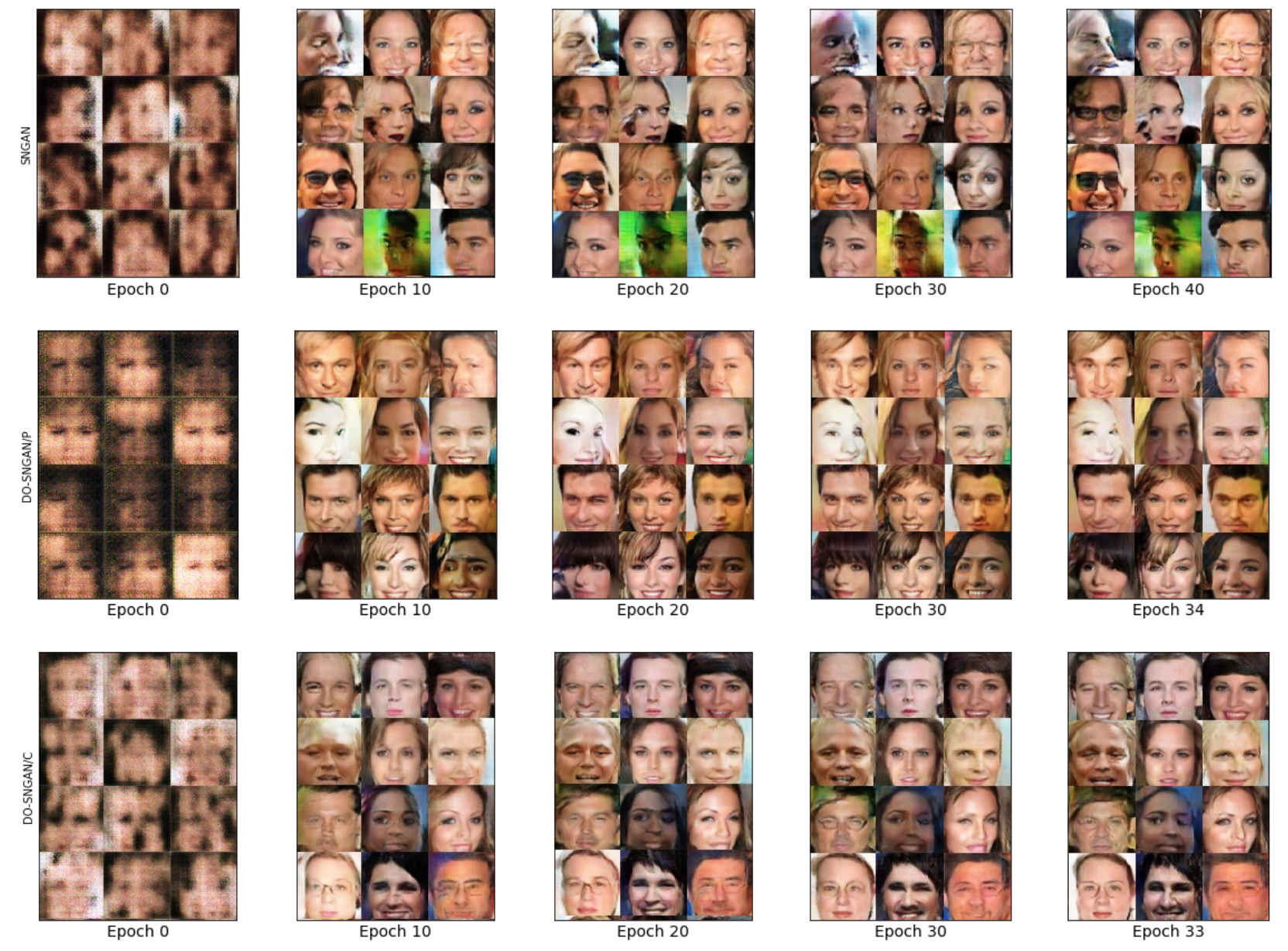}
    \caption{Training images with fixed noise for SNGAN and DO-SNGAN until termination.}
    \label{fig:celesngan}
\end{figure}

\begin{figure}[!htb]
    \centering
    \includegraphics[width=\linewidth]{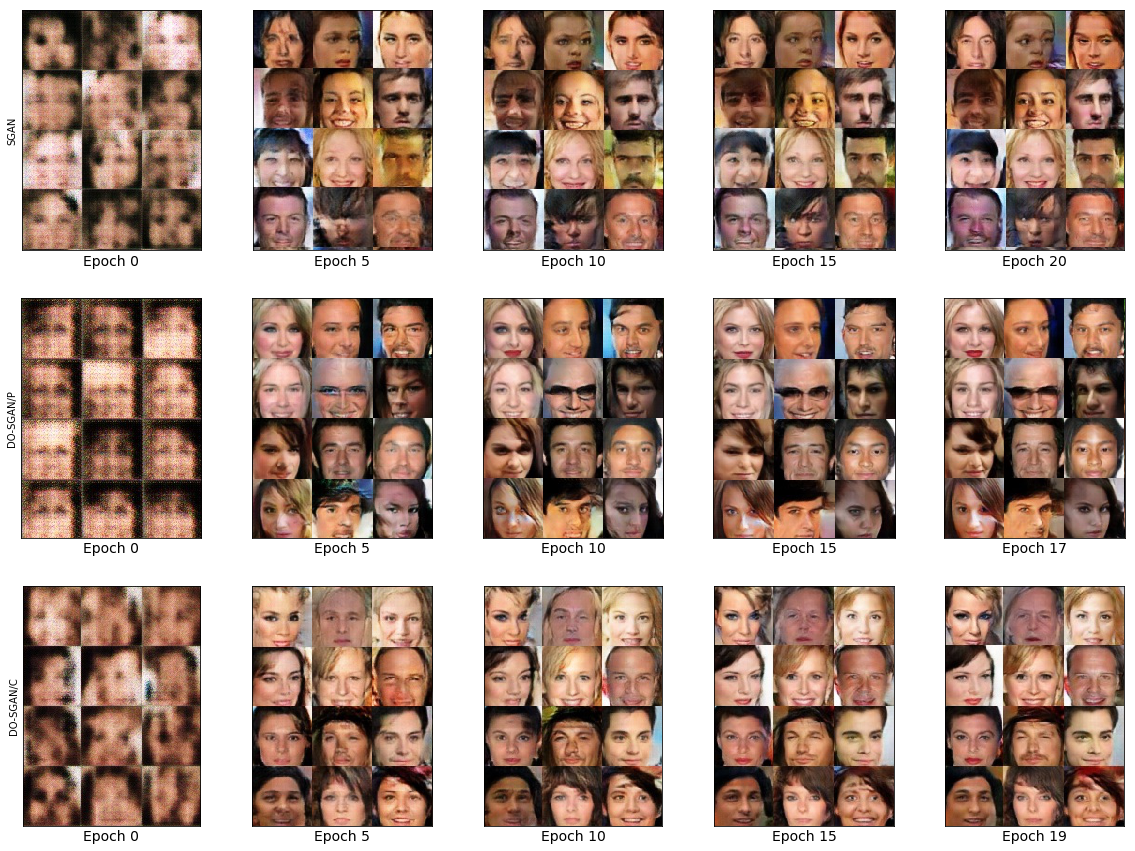}
    \caption{Training images with fixed noise for SGAN and DO-SGAN until termination.}
    \label{fig:celesgan}
\end{figure}

Figure~\ref{fig:celeb} shows the training samples of DCGAN, DO-DCGAN/P and DO-DCGAN/C through the training process. Figure~\ref{fig:celesngan} also shows those of SNGAN which is trained for 40 epochs with termination $\epsilon$ of $5 \times 10^{-5}$ for DO-SNGAN/P and DO-SNGAN/C. Figure~\ref{fig:celesgan} shows those of SGAN which is trained until 20 epochs as well as DO-SGAN/P and DO-SGAN/C with the same termination settings. The results show that DCGAN suffers from mode-collapse, generating similar face while DO-DCGAN/P can generate more plausible and varying faces. We also present the generated images of DCGAN, DO-DCGAN/P, DO-DCGAN/C, SNGAN, DO-SNGAN/P, DO-SNGAN/C, SGAN, DO-SGAN/P and DO-SGAN/C of CIFAR-10 dataset showing that the variants of DO-DCGAN, DO-SNGAN and DO-SGAN can generate better and more identifiable images than DCGAN, SNGAN and SGAN respectively. We present more of the generated samples from SNGAN, DO-SNGAN/P, DO-SNGAN/C, SGAN, DO-SGAN/P, DO-SGAN/C on CelebA dataset in Figure~\ref{fig:celebA}.

\begin{figure}[ht]
    \centering
    \begin{subfigure}[b]{.325\linewidth}
    \includegraphics[width=\linewidth]{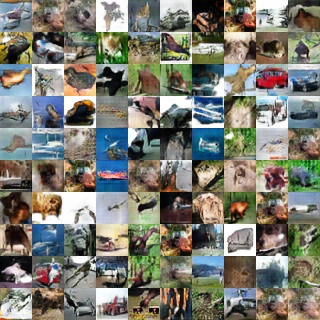}
    \caption{DCGAN}
    \end{subfigure}
    \begin{subfigure}[b]{.325\linewidth}
    \includegraphics[width=\linewidth]{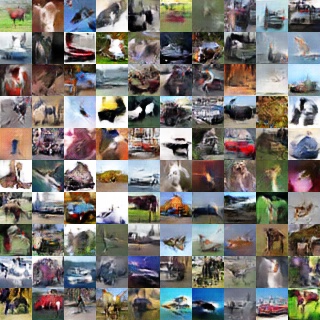}
    \caption{SNGAN}
    \end{subfigure}
    \begin{subfigure}[b]{.325\linewidth}
    \includegraphics[width=\linewidth]{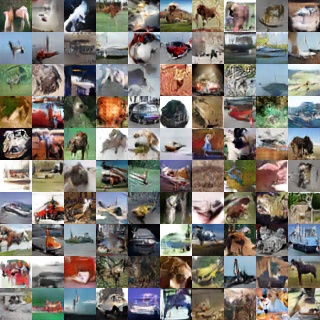}
    \caption{SGAN}
    \end{subfigure}
    \begin{subfigure}[b]{.325\linewidth}
    \includegraphics[width=\linewidth]{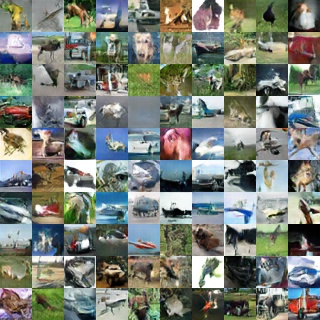}
    \caption{DO-DCGAN/P}
    \end{subfigure}
    \begin{subfigure}[b]{.325\linewidth}
    \includegraphics[width=\linewidth]{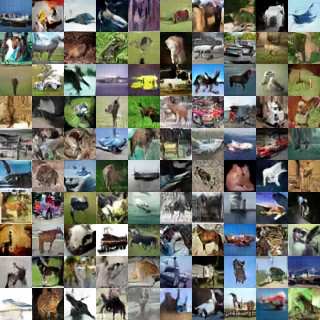}
    \caption{DO-SNGAN/P}
    \end{subfigure}
    \begin{subfigure}[b]{.325\linewidth}
    \includegraphics[width=\linewidth]{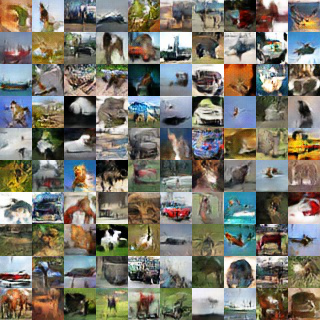}
    \caption{DO-SGAN/P}
    \end{subfigure}
    \begin{subfigure}[b]{.325\linewidth}
    \includegraphics[width=\linewidth]{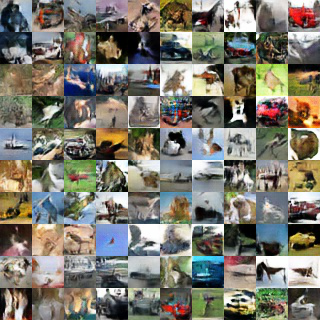}
    \caption{DO-DCGAN/C}
    \end{subfigure}
    \begin{subfigure}[b]{.325\linewidth}
    \includegraphics[width=\linewidth]{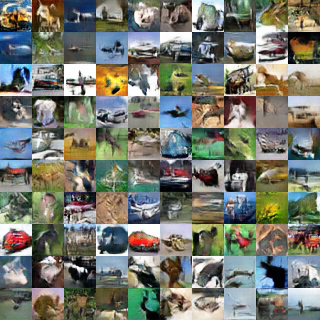}
    \caption{DO-SNGAN/C}
    \end{subfigure}
    \begin{subfigure}[b]{.325\linewidth}
    \includegraphics[width=\linewidth]{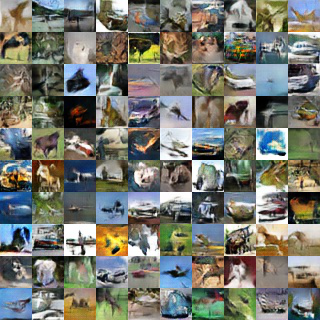}
    \caption{DO-SGAN/C}
    \end{subfigure}
    \caption{Generated images of CIFAR-10 dataset}
    \label{fig:cifar}
\end{figure}

\begin{figure}[ht]
    \centering
    \begin{subfigure}[b]{.33\linewidth}
    \includegraphics[width=\linewidth]{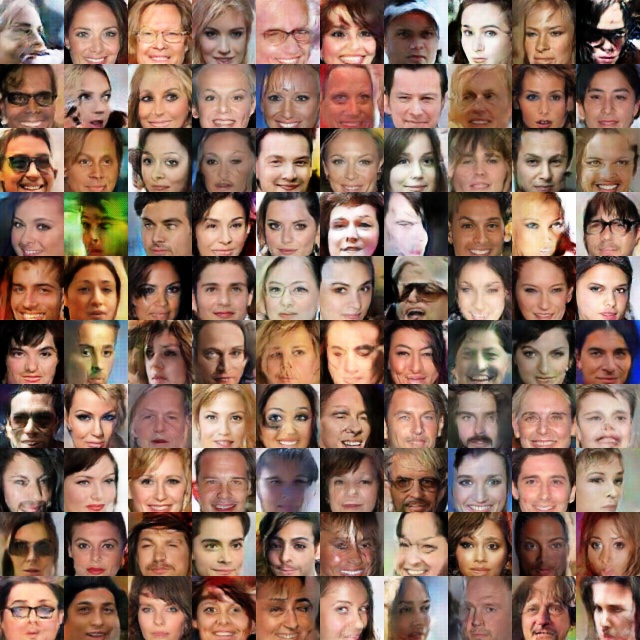}
    \caption{SNGAN}
    \end{subfigure}
    \begin{subfigure}[b]{.33\linewidth}
    \includegraphics[width=\linewidth]{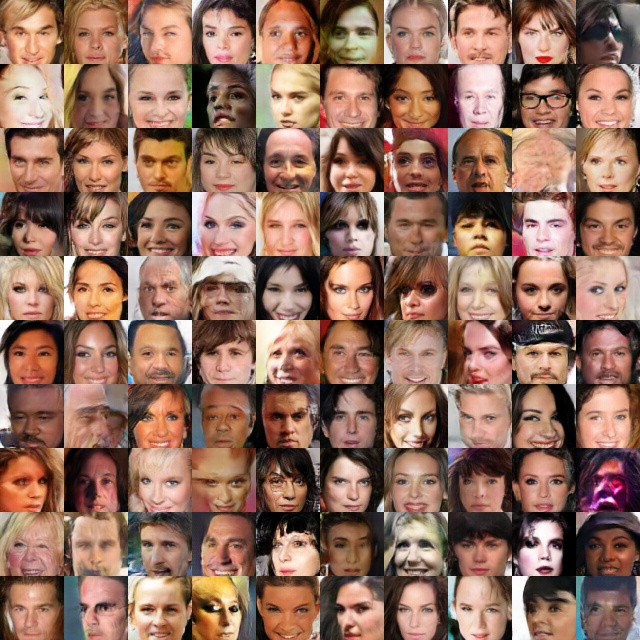}
    \caption{DO-SNGAN/P}
    \end{subfigure}
    \begin{subfigure}[b]{.33\linewidth}
    \includegraphics[width=\linewidth]{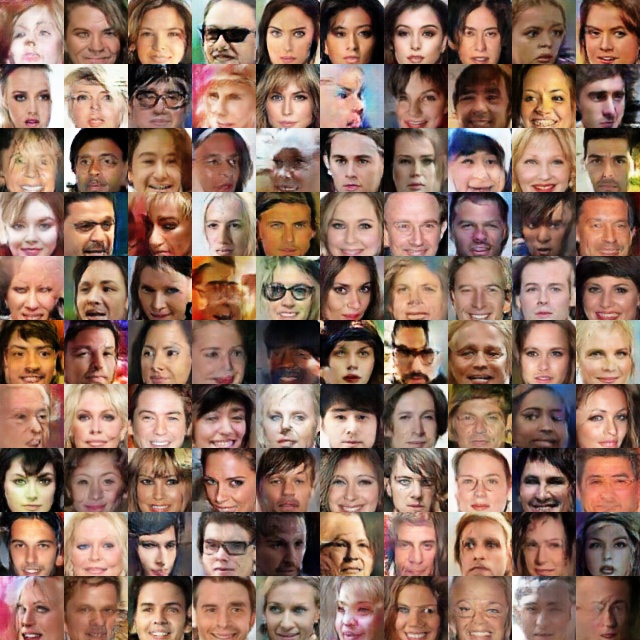}
    \caption{DO-SNGAN/C}
    \end{subfigure}
    \caption{Generated images of CelebA dataset for DO-SNGAN/P, DO-SNGAN/C and SNGAN}
\end{figure}
\begin{figure}[ht]
  \ContinuedFloat
    \centering
    \begin{subfigure}[b]{.33\linewidth}
    \includegraphics[width=\linewidth]{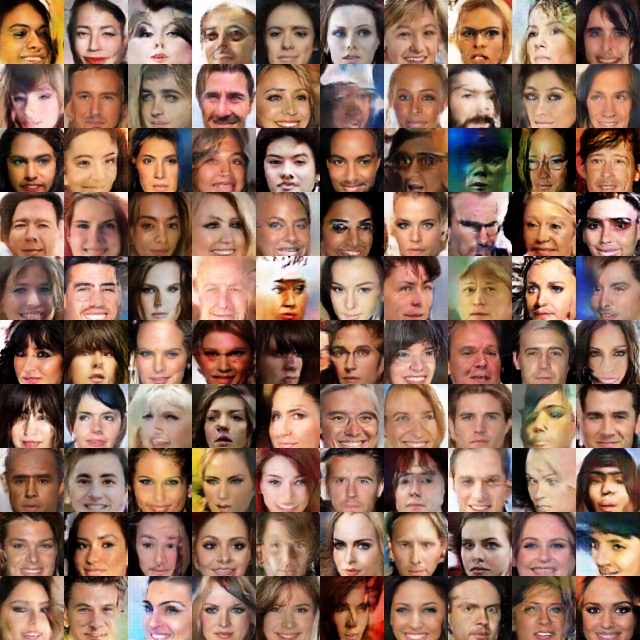}
    \caption{SGAN}
    \end{subfigure}
    \begin{subfigure}[b]{.33\linewidth}
    \includegraphics[width=\linewidth]{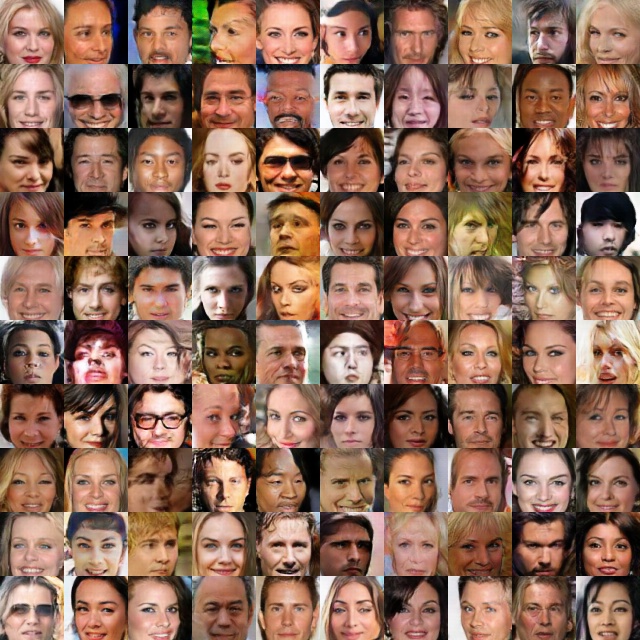}
    \caption{DO-SGAN/P}
    \end{subfigure}
    \begin{subfigure}[b]{.33\linewidth}
    \includegraphics[width=\linewidth]{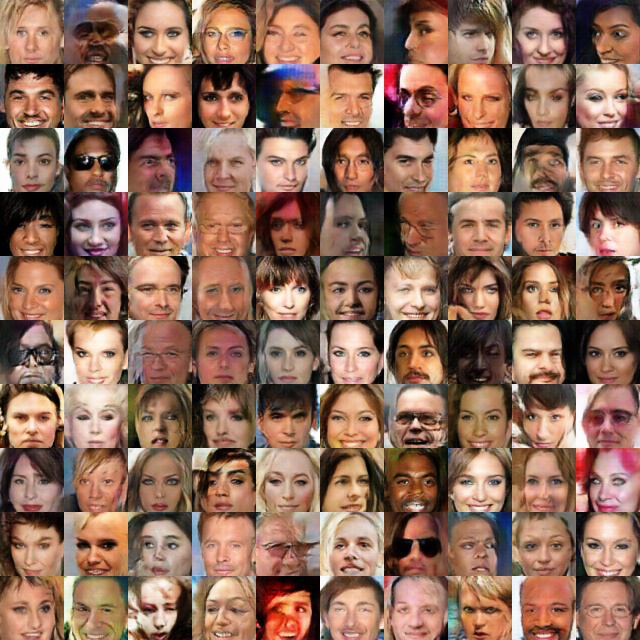}
    \caption{DO-SGAN/P}
    \end{subfigure}
     \caption{Generated images of CelebA dataset for DO-SGAN and SGAN}
    \label{fig:celebA}
\end{figure}

\section{FID score against iterations}
\label{appendixF}
To compute FID score, we use Inception\_v3 model with max pool of $192$ dimensions and the last layer as coding layer as mentioned in~\cite{heusel2017gans}. We resized MNIST, CIFAR-10 generated and test images to $32 \times 32$ and CelebA images to $64 \times 64$. The FID score against training epochs for CIFAR-10 dataset is as follows:

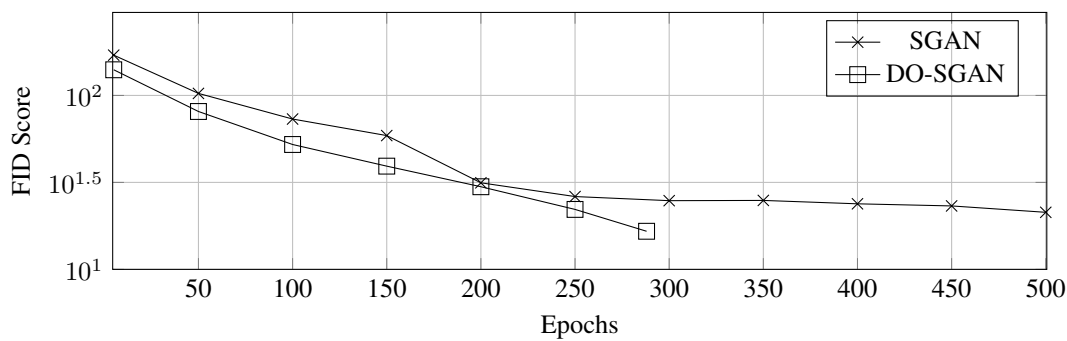
\begin{figure}[ht]
\centering
\begin{tikzpicture}[scale=1]
    \begin{axis}[ 
    legend pos= north west,
    legend style={fill=none, at=({0,1})},
    grid=major,
    legend columns=1,
    legend pos=north east,
    width= 14cm,
    height = 5cm,
    xtick={0,50,100,...,500},
    xticklabels={0,50,100,150,200,250,300,350,400,450,500},
    enlarge x limits={abs=0.55},
    ymax=300,
    ymin=10,
    ymode=log,
        xlabel=Epochs,
        ylabel=FID Score
    ]
    \addplot[color=black, mark=x,  mark size=3pt] coordinates {
        (5,170.447)
        (50,102.722)
        (100,73.174)
        (150,58.78)
        (200,31.456)
        (250,26.22)
        (300,24.83)
        (350,24.91)
        (400,23.817)
        (450,23.160)
        (500,21.284)
    };
    
    \addplot[color=black, mark=square,  mark size=3pt] coordinates {
        (5,140.447)
        (50,80.843)
        (100,52.230)
        (150,39.221)
        (200,29.862)
        (250,22.116)
        (288,16.56)
    };
    \addlegendentry{SGAN}
    \addlegendentry{DO-SGAN}
    \end{axis}
\end{tikzpicture}
\caption{FID score vs. Epochs for SGAN and DO-SGAN trained on CIFAR-10}
\label{fidscore}
\end{figure}

Figure~\ref{fidscore} presents the FID score against each epoch of training for SGAN and DO-SGAN/P on CIFAR-10. While both perform relatively well in generating plausible images, we can see that DO-SGAN/P terminates early at epoch $288$ and has a better FID score of $16.56$ compared to $24.83$ at $300$ epoch until $21.284$ at $500$ epoch for the training of SGAN. 

\newpage
\section{Choice of GAN Architectures for Experiments}
\label{appendixG}
\begin{figure}[ht]
    \centering
    \includegraphics[width=\linewidth]{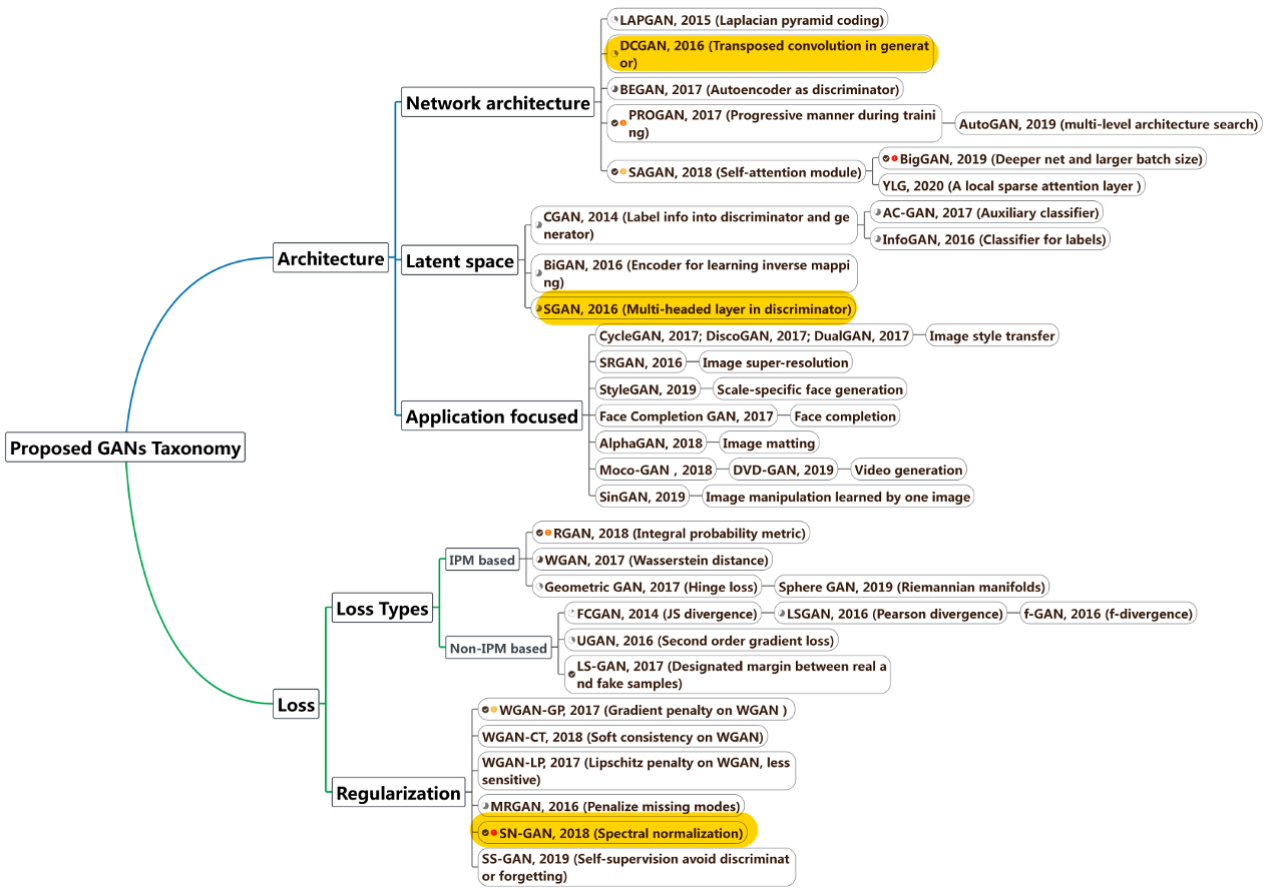}
    \caption{Taxonomy of GAN Architectures from~\cite{wang2019generative}}
    \label{gantaxo}
\end{figure}
We carried out experiments with the variants of GANs to evaluate the performance of our DO-GAN framework. We refer to the taxonomy of GANs~\cite{wang2019generative} and choose each architecture from the groups of GANs focused on Network Architecture, Latent Space and Loss: DCGAN, SNGAN and SGAN as shown in Figure~\ref{gantaxo}. We have also included comparisons with mixture architectures such as MIXGAN and MGAN.

\section{Example of Meta-matrix of DO-SGAN/P on CIFAR-10}
\begin{figure}[ht]
\centering
\begin{subfigure}[b]{.6\linewidth}
\includegraphics[width=\linewidth]{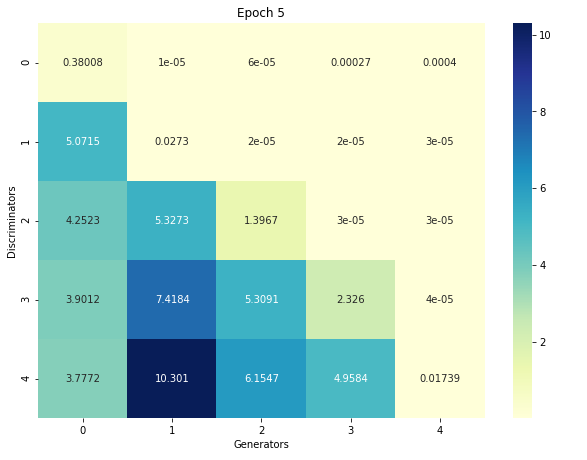}
\caption{Meta-matrix at epoch $t=5$ \\ \textbf{Meta-Strategies:} $\sigma^{5*}_{g}= [0, 0, 0, 0, 1]$, $\sigma^{5*}_{d}=[0, 0, 0, 0, 1]$ \\ \textbf{Expected Payoff:} $U^{5}(\sigma^{5*}_{g}, \sigma^{5*}_{d}) = 0.017$}
\end{subfigure}
\begin{subfigure}[b]{.6\linewidth}
\includegraphics[width=\linewidth]{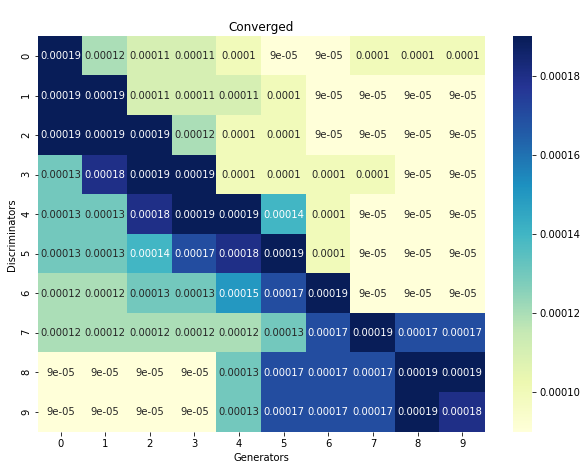}
\caption{Meta-matrix at convergence \\ \textbf{Meta-Strategies:}\\ $\sigma^{288*}_{g}= [0.015196, 0.025942,0.141215, 0.000000, 0.0989, 0.163399, 0.000000, 0.535331, 0.000000, 0.020017]$ \\ $\sigma^{288*}_{d}= [0.320082,0.004844,0.000000,0.111335,0.141493,0.071708,0.046,0.000000,0.000000,0.304538]$ \\ \textbf{Expected Payoff:} $U^{288}(\sigma^{288*}_{g}, \sigma^{288*}_{d}) = -0.000133$}
\end{subfigure}
\end{figure}

\end{document}